\documentclass[twoside]{article}
\usepackage[accepted]{aistats2024}
\usepackage[ruled,vlined]{algorithm2e}
\usepackage{wrapfig}
\usepackage{tikz}
\usetikzlibrary{arrows.meta} 
\usepackage{xcolor}

\usepackage{amsmath,amssymb,amsthm,amsfonts,latexsym,bbm,xspace,graphicx,float,mathtools,mathdots}
\usepackage{braket,caption,subcaption,ellipsis,xcolor,textcomp}
\usepackage{combelow} 
\usepackage{hyperref}
\usepackage[nameinlink]{cleveref}
\crefname{ineq}{inequality}{inequalities}
\creflabelformat{ineq}{#2{\upshape(#1)}#3}

\usepackage[letterpaper,margin=1in]{geometry}
\usepackage{enumitem} 

\newtheorem{theorem}{Theorem}

\newtheorem*{namedtheorem}{\theoremname}
\newcommand{\theoremname}{testing}

\usepackage{chngcntr}
\counterwithin{theorem}{section}

\newtheorem{proposition}[theorem]{Proposition}

\theoremstyle{definition}
\newtheorem{definition}[theorem]{Definition}
\newtheorem{remark}[theorem]{Remark}

\newtheorem{example}[theorem]{Example}

\newcommand{\ignore}[1]{}

\usepackage{xcolor}
\newcommand{\defined}{\coloneqq}
\usepackage{autonum}

\usepackage{natbib}
\setcitestyle{authoryear,open={(},close={)}} 

\newcommand{\lp}{\left(}
\newcommand{\rp}{\right)}
\newcommand{\lb}{\left[}
\newcommand{\rb}{\right]}
\newcommand{\lbr}{\left\{}
\newcommand{\rbr}{\right\}}
\newcommand{\lv}{\left\lvert}
\newcommand{\rv}{\right \rvert}

\newcommand{\thetahat}{\widehat{\theta}}

\newcommand{\convas}{\stackrel{a.s.}{\longrightarrow}}
\newcommand{\geqas}{\stackrel{a.s.}{>}}

\newcommand{\reals}{\mathbb{R}}
\newcommand{\iid}{\text{i.i.d.}}

\newcommand{\ind}{\mathbbm{1}}
\newcommand{\mc}[1]{\mathcal{#1}}

\newcommand{\Tau}{\mathcal{T}}

\DeclareMathOperator*{\argmax}{arg\,max}
\DeclareMathOperator*{\argmin}{arg\,min}

\newcommand{\computeIncrement}{\texttt{ComputeScore}\xspace}
\newcommand{\updateModel}{\texttt{UpdateModel}\xspace}

\newcommand{\swap}{\Tau_{\text{swap}}} 
\newcommand{\identity}{\Tau_{\text{id}}}
 
\newcommand{\adversarial}{\Tau_{\text{adv}}}
\newcommand{\flip}{\Tau_{\text{flip}}}
\newcommand{\gtilde}{\widetilde{g}}
\newcommand{\eqdist}{\stackrel{d}{=}}

\begin{document}

%

%

\twocolumn[

\aistatstitle{Deep anytime-valid hypothesis testing}

\aistatsauthor{ Teodora Pandeva\\University of Amsterdam \And Patrick Forr\'e\\University of Amsterdam \And Aaditya Ramdas \\Carnegie Mellon University \AND Shubhanshu Shekhar  }

\aistatsaddress{   Carnegie Mellon University} ]


\begin{abstract}

    We propose a general framework for constructing powerful, sequential hypothesis tests for a large class of nonparametric testing problems. The null hypothesis for these problems is defined in an abstract form using the action of two known operators on the data distribution.
    This abstraction allows for a unified treatment of several classical tasks, such as two-sample testing, independence testing, and conditional-independence testing,  as well as modern problems, such as testing for adversarial robustness of machine learning~(ML) models. Our proposed framework has the following advantages over classical batch tests: 1) it continuously monitors online data streams and efficiently aggregates evidence against the null, 2) it provides tight control over the type I error without the need for multiple testing correction, 3) it adapts the sample size requirement to the unknown hardness of the problem. We develop a principled approach of leveraging the representation capability of ML models within the \emph{testing-by-betting} framework, a game-theoretic approach for designing sequential tests. Empirical results on synthetic and real-world datasets demonstrate that tests instantiated using our general framework are competitive against specialized baselines on several tasks. 
\end{abstract}

\section{Introduction}
\label{sec:intro}
    We consider an abstract class of nonparametric hypothesis testing problems characterized by the action of two known operators~(denoted by $\Tau_1$ and $\Tau_2$) on the data generating distribution. Under the null, it is assumed that the transformed distributions resulting from the action of these two operators are the same, while under the alternative, it is assumed that the two transformed distributions are different. Formally, suppose the observations $Z_1, Z_2, \ldots$ are $\mc{Z}$-valued observations drawn \iid from $P_Z$, and $\Tau_i: \mc{Z} \to \mc{W}$~(for some space $\mc{W}$ possibly different from $\mc{Z}$) for $i=1, 2$ denote a pair of operators acting on the observation space. Then, we want to test 
    \begin{align}
        \hspace{-0.5em} H_0: \,  \Tau_1(Z) \stackrel{d}{=} \Tau_2(Z) \quad\text{vs.}\quad H_1:\, \Tau_1(Z) \stackrel{d}{\neq} \Tau_2(Z). \label{eq:abstract-problem-def}
    \end{align}
This abstract formulation allows for a unified treatment of several classical and modern problems ranging from two-sample and independence testing, to certifying adversarial robustness and group fairness of machine learning models (details in ~\Cref{sec:motivating-applications}). 
    
   In this paper, we propose a deep learning based strategy for designing powerful sequential tests for~\eqref{eq:abstract-problem-def}. Compared to classical batch tests, well-designed sequential tests have several benefits: they are valid under optional continuation, can consolidate evidence from possibly dependent experiments, and are computationally cheaper than permutation tests. Our results further enhance these advantages by presenting a principled approach to harness the representational power of deep neural networks (DNNs) in the context of sequential testing. As a result, our tests are particularly well-suited for handling complex data types, such as images and videos.
    
    \paragraph{Contributions.} Our main contribution is a unified data-driven strategy for designing powerful sequential tests for~\eqref{eq:abstract-problem-def}. Since this abstract formulation models a large class of practical applications, our approach effectively yields new sequential tests for all these problems in one shot. This contrasts with some recent works in this area, that propose specialized sequential tests for these individual problems.
    
    Our design strategy is guided by the principle of ``\emph{testing by betting}''~\citep{shafer2021testing}.
    This principle translates the task of designing sequential tests into that of increasing the wealth of a fictitious bettor in repeated betting games that are fair under the null. Some recent works using this principle~(details in~\Cref{sec:related-work}) decouple this task~(of setting up and betting on fair games) into separate \emph{betting} and \emph{payoff-design} problems, mainly due to analytical tractability.  
    Our work is motivated by the observation that this decoupling is unnecessary, and we instead develop a class of tests based on joint learning of both the payoff and the bets using deep learning models. In other words, unlike related sequential tests, the deep learning models in our framework are trained to directly optimize the growth rate of the wealth of the bettor, without the separation into betting and payoff design.

    Building on this basic idea,  we develop a general sequential test for the abstract testing problem in~\Cref{sec:proposed-approach}. This test relies on incrementally updated DNN~(or more generally, any machine learning) models on batches of observations. We show in~\Cref{prop:sequential-test} that this test provides tight non-asymptotic type-I-error control under the null, and is consistent~(i.e., rejects the null almost surely) against arbitrary fixed alternatives, under very mild conditions on the learning algorithm.
    
    Finally, in~\Cref{sec:exp}, we instantiate and empirically evaluate our general test for several important applications, such as two-sample testing, conditional independence testing, group invariance testing and certifying adversarial robustness. Our empirical results show that the proposed framework offers tests that are competitive and often superior to state-of-the-art tests tailored to the specific tasks.

\section{Motivating Applications}
\label{sec:motivating-applications}
 
    We now illustrate the utility of studying the abstract  testing problem~\eqref{eq:abstract-problem-def}, by showing that it models several important applications in a unified manner. 

    \begin{example}[Two-sample testing]
        \label{example:two-sample-testing}
        Given a stream of paired observations: $\{(X_t, Y_t): t \geq 1\}$ drawn \iid from a distribution $P_{X}\times P_Y$ on a product space $\mc{X} \times \mc{X}$, our goal is to decide between the null, $H_0: P_{X} = P_Y$, against the alternative ${H_1: P_X \neq P_Y}$. This is a nonparametric testing problem with a composite null and a composite alternative. The null hypothesis class, however, has an interesting symmetry: the joint distribution of $(X, Y)$ is the same as the joint distribution of $(Y, X)$.  We can formally state this as $H_0: (X,Y) \eqdist \swap((X,Y))$, where $\swap: \mc{X} \times \mc{X} \to \mc{X} \times \mc{X}$, and $\swap((x,y)) = (y, x)$. 
    \end{example}

    \begin{example}[Conditional independence testing]
        \label{example:conditional-independence}
        Given observations $\{(U_t, V_t, W_t):  t \geq 1\}$ drawn i.i.d. from $P_{UVW}$, we want to test whether ${U\perp \!\!\! \perp  V| W}$ or not. This problem is fundamentally impossible without further assumptions~\citep{shah2020hardness}, and a  common structural assumption is to that the conditional $P_{U|W}$ is known~(the \emph{model-X} assumption \citep{candes2018panning}). We can now reframe this problem as follows: 
        \begin{itemize}[leftmargin=*]
            \item Given $(U, V, W)$, generate a new $\widetilde{U} \sim P_{U|W}(\cdot |W)$, and let $Z$ denote   $\big( (U, V, W), (\widetilde{U}, V, W) \big)$.  
            \item Let $\Tau_1$ and $\Tau_2$ denote the coordinate projections:  $\Tau_1(Z) = (U, V, W)$ and $\Tau_2(Z) = (\widetilde{U}, V, W)$. 
        \end{itemize}
        With these definitions, conditional independence~(CI) testing falls under the framework of~\eqref{eq:abstract-problem-def}.
    \end{example}
    Testing for invariance under group actions, such as rotations, is another instantiation of~\eqref{eq:abstract-problem-def}. 
        \begin{example}[Rotation invariance testing]
        \label{example:group-invariance}
         Given a stream of observations $\{(X_t, Y_t): t \geq 1\}$, where the $X_t$'s denote of images of the (handwritten) digit ``$6$'', while dataset $Y_t$'s are images that, at a glance, represent the digit ``$9$''. However, these may essentially be the digit ``$6$'' but rotated. We aim to determine the statistical relationship between $X_t$ and $\Tau_{180}(Y_t)$: the  $180$ degree rotations of $Y_t$. Essentially, we want to decide whether $(Y_t)_{t \geq 1}$ are merely rotated versions of ``$6$'', or truly represent the digit ``$9$''. 
        Using the swap operator from~\Cref{example:two-sample-testing},   we define two distinct operators: $ \Tau_1 = (\Tau_{180}, \Tau_{\text{id}})\circ \Tau_{\text{swap}}$, and $\Tau_2 = (\Tau_{\text{id}}, \Tau_{180})$, with $\Tau_{\text{id}}$ being the identity mapping. Then, the above-defined test is equivalent to testing  $ H_0: \Tau_1(Z) \stackrel{d}{=} \Tau_2(Z) $, where $Z = (X,Y)$.
    \end{example}
    Our general framework is not restricted to simple operators with closed-form expressions as in the examples above. In fact, the operators involved can even be general function approximators, such as large neural networks. 
    \begin{example}[Adversarial Examples]
        \label{example:adversarial-attacks}
        We now consider the problem of certifying the robustness of a trained machine learning model $h$ to adversarial perturbations~\citep{szegedy2014intriguing}. In particular, let $\adversarial$ denote the adversarial attack which maps an input $X$ to its adversarially perturbed version $\widetilde{X}$. Furthermore,  let $\Tau_h$ denote the output of a specific layer~(for example, a bottleneck layer) of the model. 
        Then, our goal is to decide if the distributions of $Y = \Tau_h(X)$ and $\widetilde{Y} = \Tau_h(\widetilde{X})$ are equal or not.  In other words, the null states that the distribution of $X$ after applying $\Tau_h$ and $\Tau_h \circ \adversarial$ is the same. 
    \end{example}

 Further examples of ~\eqref{eq:abstract-problem-def}, such as tests for group fairness and independence, are available in~\Cref{app:more-examples}.

\section{Related Work}
\label{sec:related-work}
    The abstract testing problem studied in this paper is motivated by the  general task of testing for invariance to the action of finite groups, stated as the  \emph{randomization hypothesis} by~\citet[Definition~17.2.1]{lehmann1986testing}.    In fact,  the ideas we develop can also be extended to deal with the formulation of \citet{lehmann1986testing}~(See \Cref{sec:rand-test}).  

    From a methodological perspective, our techniques are related to the growing body of recent work on  safe anytime-valid inference~(SAVI), surveyed by~\citet{ramdas2022game}. Our design strategy  follows the principle of \emph{testing by betting}~\citep{shafer2021testing}, which states that the evidence against a null can be precisely characterized in terms of the gain in wealth of a (fictitious) bettor, who repeatedly bets on the observations in betting games with  odds that are fair (or sub-fair) under $H_0$. 
    This principle has been used by several authors, such as~\citet{shekhar2023nonparametric, podkopaev2023skit, podkopaev2023predictive, shaer2023model}, to transform different hypothesis testing problems into that of designing relevant betting strategies and payoff functions. For example,~\citet{shekhar2023nonparametric} considered the two-sample testing problem~(\Cref{example:two-sample-testing}), 
    and defined the wealth process of the bettor  as follows for $t \geq 1$: 
    \begin{align}
        &W_t = W_{t-1} \times \lp 1 + \lambda_t  \big(g_t(X_t) - g_t(Y_t)  \big) \rp,
    \end{align}
    for some $[-1/2, 1/2]$-valued payoff functions $(g_t)_{t \geq 1}$, and for bets $\lambda_t \in [-1, 1]$ and $W_0=1$. By construction, the process $(W_t)_{t \geq 0}$ satisfies the requirement of fair payoffs under the null, as it is a non-negative  martingale.  The term $\lambda_t$ is a predictable bet, whose absolute value denotes the fraction of the accumulated wealth that is placed at stake in round $t$. Hence, the approach of~\citet{shekhar2023nonparametric}, reduces the problem of two-sample testing into that of developing appropriate strategies for selecting $(g_t)_{t \geq 1}$~(the \emph{prediction strategy}) and $(\lambda_t)_{t \geq 1}$~(the \emph{betting strategy}). 
    There exist off-the-shelf betting strategies in the literature on online learning, such as the online Newton step~(ONS) strategy~\citep{hazan2007logarithmic}, that ensure an exponential growth of the wealth process for arbitrary $(g_t)_{t \geq 1}$ under the alternative. For the prediction strategy,~\citet{shekhar2023nonparametric} suggested selecting $(g_t)_{t \geq 1}$ that approximate the \emph{witness function}~($g^*$) associated with statistical distance metrics with variational representations~(such as Kolmogorov-Smirnov metric, kernel MMD, $f$-divergence, etc). A similar approach was followed by~\citet{podkopaev2023skit} and~\citet{shaer2023model} for the problems of independence and conditional independence testing. 
  
    Machine learning models (classifiers or regressors) have proven highly effective in developing  tests for complex data structures; see~\citet{kim2021classification} and references therein for more details. Even within the SAVI framework, some prior works such as \citet{podkopaev2023predictive, pandeva2022valuating, lheritier2018sequential} propose two-sample and independence tests based on classifiers.  Unlike these works, our approach is geared towards a general class of problems modeled by~\eqref{eq:abstract-problem-def}, and uses models trained directly to optimize the test performance.

\section{Methodology}
\label{sec:proposed-approach}
   We study the testing problem described in~\eqref{eq:abstract-problem-def}, under the assumption that we observe a stream of datapoints arriving in mini-batches $(B_t)_{t \geq 1}$, where $B_t = \{Z_{(t-1)b+1}, \ldots, Z_{tb}\}$ denotes the $t$-{th} mini-batch consisting of $b$ \iid observations. Our objective is to design a procedure to continuously monitor the stream, aggregate evidence against the null, and stop and reject the null as soon as sufficient evidence is collected.  Formally, such procedures are called \emph{sequential tests of power one}, following~\citet{darling1968some}, and we state their definition below.
    \begin{definition}
        \label{def:power-one-sequential-test} Given a significance level $\alpha \in (0, 1)$, and a stream of mini-batches, $\{B_t: t \geq 1\}$, consisting of samples drawn i.i.d. from a distribution $P_Z$, consider the testing problem introduced in~\eqref{eq:abstract-problem-def}. A level-$\alpha$ sequential test of power one for this problem is a stopping time, $\gamma$, adapted to the natural filtration $(\mc{F}_t)_{t \geq 0}$, with $\mc{F}_t = \sigma(B_1, \ldots, B_t)$, satisfying 
        \begin{align}
            \mathbb{P}_{H_0} \lp \gamma < \infty \rp \leq \alpha, \; \text{and} \; 
            \mathbb{P}_{H_1}\lp \gamma < \infty \rp = 1. \label{eq:seq-test-def}
        \end{align}
        In other words, $\gamma$ denotes a data-driven stopping rule at which the data-analyst stops collecting more data, and rejects the null. It is required that  if the null holds, the probability that the test stops, i.e. it rejects the null, is bounded by $\alpha$. In contrast, if the alternative is true this probability should be 1 which guarantees the consistency of the test.
    \end{definition}

    \paragraph{Oracle Sequential Test.}  We begin by formulating an `oracle' sequential test for~\eqref{eq:abstract-problem-def}, that assumes full knowledge of the true distribution $P_Z$. While this test is impractical, it provides the template for designing our practical data-driven sequential test. 
    
    Let $\mc{G} = \{g_\theta: \theta \in \Theta \}$ denote a class of  machine learning models parameterized by $\Theta$. For example, $\mc{G}$ might represent a class of deep learning models, with the parameter set $\Theta$ specifying the architecture. Furthermore, assume that all the functions $g_\theta$ are $[-q, q]$-valued for some $q \in (0, 1/2)$,
    and for all choices of $\theta \in \Theta$. This is easily ensured in DNNs by applying a sigmoid function to the output of the last layer.
    Since DNNs satisfy the universal approximation property  for sufficiently large choices of the architecture \citep{hornik1991approximation,cybenko1989approximation}, for  two distinct distributions $P \neq Q$ on $\mc{W}$, we can infer that
    \begin{align}
        \sup_{\theta \in \Theta} \mathbb{E}_P[g_\theta(W)] - \mathbb{E}_Q[g_\theta(W)] > 0. \label{eq:sup-theta-1}
    \end{align}
    
    A simple consequence of~\eqref{eq:sup-theta-1} is that $\Tau_1(Z)$ and $\Tau_2(Z)$ have different distributions, if and only if $\sup_{\theta \in \Theta} \mathbb{E} \lb \log \lp 1+  \gtilde_{\theta}(Z, \Tau_1, \Tau_2)\rp  \rb > 0$, where $\gtilde_{\theta}(z, \Tau_1, \Tau_2) \defined g_{\theta}\circ \Tau_1(z) - g_{\theta}\circ \Tau_2(z)$.

    In the sequel, we will drop the $\Tau_1$ and $\Tau_2$ dependence of $\gtilde$, and simply write $\gtilde_{\theta}(z) \equiv \gtilde_{\theta}(z, \Tau_1, \Tau_2)$. 
    
    Using the above observation, we can define the `oracle' parameter, $\theta^* \equiv \theta^*(P_Z, \Tau_1, \Tau_2)$ as follows: 
    \begin{align}
        \theta^*  \in \argmax_{\theta \in \Theta}  \; \mathbb{E}_{P_Z} \lb \log \lp 1+ \gtilde_{\theta}(Z) \rp \rb. \label{eq:theta-star}
    \end{align}
    Thus, $\theta^*$ represents the $\log$-optimal function in $\mc{G}$, and we  can use it to define an \emph{oracle sequential test} 
    \begin{align}
        &\gamma^* = \inf \{t \geq 1: W^*_t \geq 1/\alpha \},  \label{eq:oracle-test-def}
    \end{align}
    where $W^*_t = \prod_{l=1}^t \prod_{Z \in B_l} \lp 1 + \gtilde_{\theta^*}(Z) \rp$.
    It is easy to verify that $\gamma^*$ is a sequential test according to~\Cref{def:power-one-sequential-test}. In particular, it ensures the control of type-I error at level-$\alpha$ under $H_0$, and is finite almost surely under the alternative. 

    Clearly, the test defined above is not practical, as it depends on the `oracle' parameter $\theta^*$, which is a function of  $P_Z$. To construct a practical test, we instead use predictable empirical estimates of $\theta^*$\footnote{As a warm-up, we also construct and theoretically analyze a practical batch test based on sample-splitting  in~\Cref{sec:batch-test-sample-splitting} of the appendix. }. This is explained in detail in the following.
    
    \paragraph{Practical sequential test.}
    For the practical test, we propose to replace $(W_t^*)_{t \geq 0}$ in~\eqref{eq:oracle-test-def} with a data-driven process $(W_t)_{t \geq 0}$, that we refer to as the \emph{wealth process} following  the standard convention as discussed in~\Cref{sec:related-work}. We set $W_0=1$, denoting the bettor's initial investment, and update $W_t$ to $W_{t-1}\times S_t$ for $t \geq 1$, with $S_t$ representing the gain~(or loss) made  while betting on the $t$-{th} batch of observations. We refer to this increment $S_t$ as the \emph{betting score} following~\citet{shafer2021testing}. 
        
    \Cref{algo:sequential-test} provides a detailed pseudocode describing the steps involved in the construction of our sequential test. The inputs to this algorithm include the stream of mini-batches~$(B_t)_{t \geq 1}$,  test-specific operators~($\Tau_1$ and $\Tau_2$), significance level $\alpha \in (0,1)$, the maximum time horizon $T_{\max}$, and a deep learning~(or any other machine learning) model initialized at $\theta_0$. The algorithm then proceeds by repeating the following steps for all $t \geq 1$: it observes the next mini-batch $B_t$, computes the betting score $S_t$ by calling the \computeIncrement subroutine, and updates the model to $\theta_t$ by calling the \updateModel subroutine. The updated wealth $W_t$ is obtained by using the betting score $S_t$, and the algorithm stops and rejects the null if $W_t$ exceeds the threshold $1/\alpha$. 
    
    To complete the description of our scheme, we now present the details of the two subroutines. 
    
    \textbf{Compute Betting Score (\computeIncrement).}  In round $t \geq 1$, this subroutine takes the inputs: \vspace{0.25em} \\
        $\bullet\;$ $\mc{D}_{t-1} = \cup_{i=1}^{t-1} B_{i}$: the data observed so far. \\
        $\bullet\;$ The model $\theta_{t-1}$, trained on $\mc{D}_{t-1}$. \\
        $\bullet\;$ $B_t = \{Z_{(t-1)b+1}, \ldots, Z_{tb}\}$: the new mini-batch. \\
        $\bullet\;$  $\Tau_1, \Tau_2$: the operators defining the null.\vspace{0.25em} \\
    Using these inputs, this subroutine computes and returns the next multiplicative increment, or betting score, ($S_t$) of the wealth process, which is defined as 
    \begin{align}
        S_t = \prod_{j=1}^b\lp 1 +   \gtilde_{\theta_{t-1}}(Z_{(t-1)b+j}) \rp.  \label{eq:wealth-increment}
    \end{align}
    With this, we update the wealth process ${W_t \leftarrow W_{t-1} \times S_t}$. We reject the null if $W_t\ge \alpha^{-1}$, otherwise, we proceed to the next step:
        
    \textbf{Model Update (\updateModel).}   
    This subroutine updates and returns a model $\theta_{t}$ on the data set ${\mc{D}_t = \mc{D}_{t-1} \cup B_t}$, i.e. $\theta_{t}$ maximizes the objective
    \begin{align}
    \label{eq:objective}
        \theta_t\in \arg\max_{\theta \in \Theta}\sum_{l=1}^t\sum_{Z\in B_l} \log(1+\gtilde_{\theta}(Z))
    \end{align}
    The input \texttt{training\_params} refers to all training parameters needed for model refinement. In particular, this includes parameters that are required for the optimization (such as the learning rate for the optimizer) and those that set the criteria for early stopping to prevent model overfitting. Importantly, our framework is versatile enough to incorporate any learning and model selection process, including methods such as cross-validation. For implementation details of this step see \Cref{sec:exp,app:implementation}.

    The stopping time constructed by~\Cref{algo:sequential-test} can be formally defined as 
    \begin{align}
        \gamma = \inf \{t \geq 1: W_t \geq 1/\alpha\}.  \label{eq:sequential-test-def}
    \end{align}
    As mentioned earlier, $\gamma$ represents the time at which the data-analyst stops collecting more observations~(i.e., mini-batches), and declares the null to be not true. We now show that $\gamma$ is finite under the null with a probability no larger than $\alpha$, and under the alternative with probability $1$~(assuming $T_{\max}$ is large enough). In simpler terms, the following theoretical result confirms that the sequential test from \Cref{algo:sequential-test}  is consistent while maintaining non-asymptotic type I error control.
    
    \begin{proposition}
        \label{prop:sequential-test} Suppose the learning algorithm satisfies the condition 
        \begin{align}
            \liminf_{t \to \infty} \frac{\mathbb{E}[\log(1+\gtilde_{\theta_t}(Z))|\mc{F}_{t}]}{2 c \sqrt{\log (t) /t}} \geqas 1, \quad \text{under } H_1 
        \end{align}
        for a universal constant $c$. 
        Then, we have 
        \begin{align}
            \mathbb{P}_{H_0}(\gamma < \infty) \leq \alpha, \; \text{and} \; \mathbb{P}_{H_1}(\gamma < \infty) = 1. 
        \end{align}
        In words, $\gamma$ is a sequential level-$\alpha$ test of power one. 
    \end{proposition}
The proof is provided in the appendix~(\Cref{proof:sequential-test}).        
\begin{remark}
           \label{remark:batch-test} The condition required of the learning algorithm by~\Cref{prop:sequential-test} for consistency of $\gamma$ under $H_1$ is very mild. Informally, we only require $\mathbb{E}[\log(1+\gtilde_{\theta_t}(Z))|\mc{F}_t]$ to be larger than $2c \sqrt{\log (tb) /tb}$ for large $t$, and in particular, this value can even converge to $0$. In practice, most models converge to a local optimum $\theta_\infty$ with ${\mathbb{E}[\log(1+\gtilde_{\theta_\infty}(Z))] >0}$, which is much stronger than the condition required above. Such strong performance guarantees on learning algorithms can  lead to stronger statistical properties of the test (such as bounds on $\mathbb{E}_{H_1}[\gamma]$). We leave such extensions to future work.  
       \end{remark}
            \begin{algorithm}
        \caption{Sequential Test}
       \label{algo:sequential-test}
        \KwIn{$\{B_t\}_{t\ge 1}$ (batch stream), $\Tau_1, \Tau_2$ (operators),  $T_{\max}$~(maximum rounds of observations), $\alpha$~(size of the test), $\theta_0$~(a trainable model).} 
        $W_0 \leftarrow 1$, $\mc{D}_0 \leftarrow \emptyset$. \\
        Initialize the model to an arbitrary value $\theta_0$. \\
        \For{$t \leftarrow 1$ \KwTo $T_{\max}$:}{
            Observe the next batch $B_t = \{Z_{(t-1)b+1}, \ldots, Z_{tb}\}$ \; 
            Compute the multiplicative increment: $S_t \leftarrow$ \computeIncrement($ B_t,  \theta_{t-1}, \Tau_1, \Tau_2, \sigma)$ \; 
            Update the wealth process: $W_t \leftarrow W_{t-1} \times S_t$  \; 
            Check for stopping condition\; 
            \If{$W_t \geq 1/\alpha$}{
                Stop and reject the null 
                }
            Increment Data: $\mc{D}_{t} = \mc{D}_{t-1}\cup B_{t}$ \;    
            Update the model: $\theta_t \leftarrow \updateModel(\mc{D}_{t}, \theta_{t-1}, \texttt{training\_params})$\; %
        }

    \end{algorithm}
        
\section{Extension to the Randomization Hypothesis Testing}
\label{sec:rand-test}
The abstract problem~\eqref{eq:abstract-problem-def} tests whether the data distributions, after being transformed by operators $\Tau_1$ and $\Tau_2$, are the same or not. We now study a generalization of this in which $\Tau_1$ and $\Tau_2$ could be one among finite disjoint classes of  operators. This extension is motivated by the so-called randomization hypothesis assumption of~\citet[\S~17.2.1]{lehmann1986testing}.  More formally, we consider testing problems with null defined as
\begin{align}
    H_0:\; \Tau_1(Z) \stackrel{d}{=} \Tau_2(Z), \; \forall \Tau_1\in \mathcal{O}_1, \forall\Tau_2\in \mathcal{O}_2\label{eq:rand-test}
\end{align}
for finite disjoint sets of operators $\mathcal{O}_1 \subset \mathcal{W}^{\mathcal{Z}}$ and $\mathcal{O}_2 \subset \mathcal{W}^{\mathcal{Z}}$.
This formulation is highly flexible and covers a wide range of complex hypotheses, from those testing invariance under permutations to diverse group actions, as illustrated in Example~\ref{example:rotation-invariance}.
By using the general strategy we developed in~\Cref{sec:proposed-approach}, we can construct a sequential test for this problem, by simply  modifying the training objective from \eqref{eq:objective} of the machine learning model as follows: 
\begin{align}
            &\theta_t \in \argmax_{ \theta \in \theta} \; \sum_{l=1}^t\sum_{Z\in B_t}  \log \lp 1 + \mathbb{E}_{\Tau_1,\Tau_2}\left[\gtilde_{\theta}(Z, \Tau_1,\Tau_2)\right] \rp,\\
            &\mathbb{E}_{\Tau_1,\Tau_2}\left[\gtilde_{\theta}(Z, \Tau_1,\Tau_2)\right] = \frac{\sum_{\Tau_1, \Tau_2} g_{\theta}\circ \Tau_1(Z)-g_{\theta}\circ \Tau_2(Z)}{|\mathcal{O}_1||\mathcal{O}_2|}.
        \end{align}
Essentially, the model  aims to maximize the growth rate by calculating the average payoffs across all operator pairs. If there is a noticeable difference in one of the components $ g_{\theta}\circ \Tau_1(Z)-g_{\theta}\circ \Tau_2(Z)$ it suggests the alternative hypothesis might hold. Similar to the previous section, the model can be incrementally refined with each new data batch. At the same time, the wealth process can be monitored with betting scores at time $t$ defined as
\begin{figure*}[!ht]
    \centering
    \includegraphics[width=\textwidth]{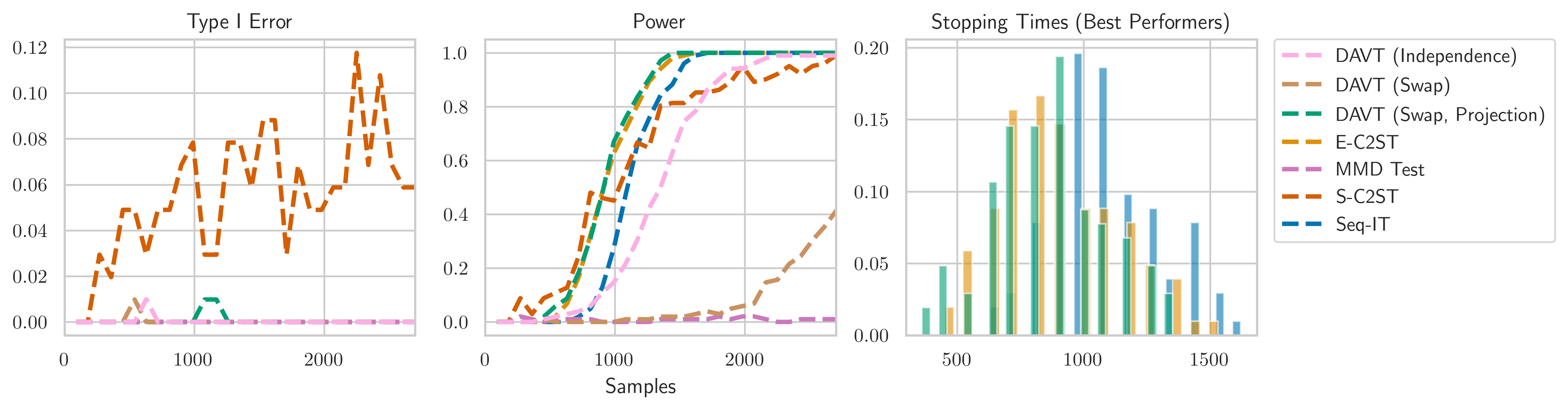}
    \caption{Power and type I error analysis for the Blob dataset.  We compared three variations of our method (DAVT (Independence), DAVT (Swap), DAVT-(Swap,Projection)) to sequential baselines (E-C2ST, Seq-IT) and non-sequential ones (S-C2ST, MMD Test). We fixed the batch size to be $90$. 
    DAVT (our method) with the swap and projection operators (in green) and E-C2ST show the best performance, followed by Seq-IT. This order in performance is also confirmed by the histogram  of the stopping times.}
    \label{fig:blob-exp}
\end{figure*}
\begin{align}
    S_t = \prod_{j=1}^b (1+\mathbb{E}_{\Tau_1,\Tau_2}\left[\gtilde_{\theta_{t-1}}(Z_{(t-1)b+j}, \Tau_1,\Tau_2)\right]). 
\end{align}
Following the same arguments as~\Cref{prop:sequential-test}, we can show that the resulting sequential test for \eqref{eq:rand-test} provides finite-sample  type I error control and consistency guarantees.

\section{Experiments}
\label{sec:exp}

 In this section, we instantiate our general strategy, which we refer to as \textbf{DAVT}, to a wide range of tasks: two-sample testing, rotation invariance testing, robustness to adversarial attacks, and conditional independence under the model-X assumption. We compare DAVT to popular nonparametric baselines, including \textit{sequential methods} such as the two-sample tests: E-C2ST \citep{lheritier2018sequential,pandeva2022valuating} and Seq-IT with a batch-wise ONS betting strategy \citep{podkopaev2023predictive} and the conditional independence test ECRT \citep{shaer2023model}, and \textit{permutation-based nonparametric tests} such as the MMD test \citep{gretton2012kernel} and the classifier two-sample test (S-C2ST) \citep{kim2021classification,lopezpaz2017revisiting}.  The latter techniques have a ``non-sequential'' decision rule based on a $p$-value computed on a single batch. For a fair comparison, we implement all considered data-driven tests using the same net architectures.  Details of the selected DNNs, and their training procedures can be found in \Cref{app:implementation}. We evaluate all tests by monitoring the rejection rates (power and type I error) over time from 100 independent runs, performed at a significance level of ${\alpha=0.05}$. In our  empirical evaluation on standard benchmark datasets, DAVT shows competitive, if not superior, results to its specialized counterparts. 

\subsection{Two-Sample Testing} 
    The two-sample testing problem can be modeled in several ways using distinct operators within our framework. 
    For instance, in~\Cref{example:two-sample-testing}, we modeled it using the operators ${\Tau_1 := \Tau_{\text{swap}} }$ and ${ \Tau_2 := \Tau_{\text{id}}}$.  Alternatively, the operators $ {\Tau_1:= \Tau_{\text{proj, swap}}= \Tau_{\text{proj}}\circ \Tau_{\text{swap}}}$ and $ {\Tau_2: = \Tau_{\text{proj}} }$ with ${ \Tau_{\text{proj}}(x,y) = x }$ also characterize the two-sample testing problem. Finally, another option is to formulate two-sample testing as an instance of independence testing. This is achieved by introducing a binary variable $L_t$ and a variable $W$ such that $P(W|L=1) = P(X)$ and $P(W|L=0) = P(Y)$. Then, the two-sample test transforms into testing the independence of $W$ and $L$, using the operator defined Example~\ref{example:independence-testing}.

 Building on this observation, we evaluate the above two-sample tests along with the baselines on the Blob dataset \citep{chwialkowski2015fast}. This dataset contains two classes of data, $X$ and $Y$, both representing nine Gaussians on a two-dimensional grid that differ in their variances (see \Cref{fig:blob}).  The results, summarized in \Cref{fig:blob-exp}, show the superior performance of DAVT (our method) when used with $\Tau_{\text{proj,swap}}$ (shown in green), closely followed by the sequential methods E-C2ST and Seq-IT. Conversely, other DAVT variants did not deliver comparable performance. For example, using only the swap operator (DAVT-Swap) results in poor test performance. This is due to the inherent attempt of the neural network to find correlations between $X$ and $Y$, which contradicts the problem setup that $X$ and $Y$ are independent. We explore  the test performance for dependent  $X$ and $Y$  in \Cref{subsub:indep}. Moreover, DAVT-Independence is not as powerful as the top sequential methods on this task, but it achieves maximum power faster than the non-sequential  methods.

Overall, our two-sample experiment highlights the strengths of sequential methods over batch tests. By construction, well-designed sequential tests continuously monitor the data stream to accumulate evidence against the null. Thus, instead of setting a fixed sample size,  using sequential methods allows for dynamically tailoring the sample size to the complexity of the task in a data-driven manner. This adaptability is illustrated in \Cref{fig:blob-exp}, showing the stopping times distribution of the top three performers from  the power experiment. Here, DAVT (Swap, Projection) and E-C2ST reject the null more quickly than Seq-IT, indicating a more efficient use of data.

\subsection{Conditional Independence Testing under Model-X Assumption}

We use the synthetic example of \citep{shaer2023model} with the following data generation model. We construct mini-batches containing 100 observations of the random variables $U, V, W$ for which we test $H_0: V \perp \!\!\! \perp U|W$. These triples come from the model $W\sim \mathcal{N}(0, I_d)$ and $U|W=w \sim \mathcal{N}(a^\top w, 1)$, where under the alternative hypothesis, we use ${V|W=w, U=u \sim \mathcal{N}((b^\top w)^2 + 3w, 1)}$, while under the null hypothesis we use ${V|W=w, U=u \sim \mathcal{N}((b^\top w)^2, 1)}$. We recall that the model-X assumption implies that at testing time $t$, we have access to the true data distribution $U|W$. 

\begin{figure}
    \centering
    \includegraphics[width=0.78\columnwidth]{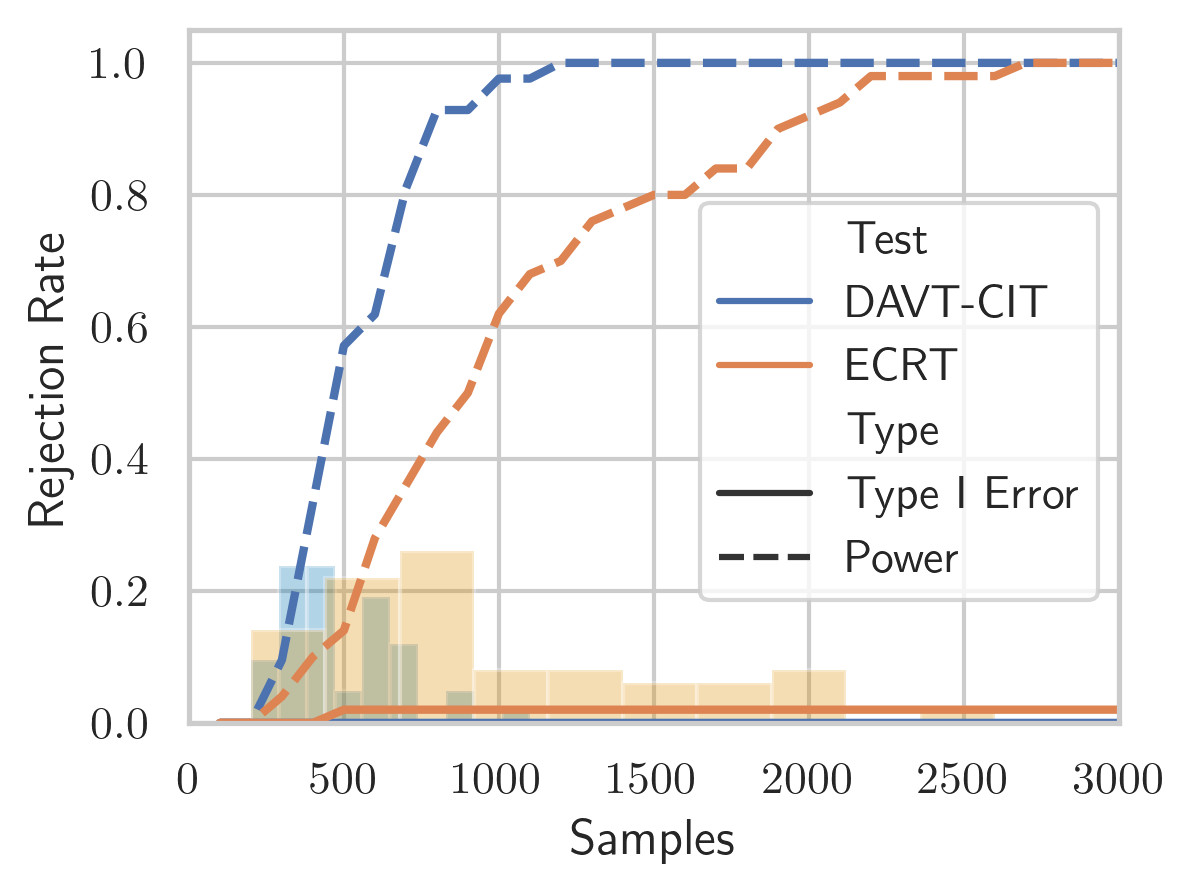}
    \caption{CIT Power and Type I error control. DAVT-CIT (ours) increases power faster than ECRT while keeping a very low Type I error.}
    \label{fig:cit}
\end{figure}

We instantiate our conditional independence test~(called the DAVT-CI) using the operators defined in~\Cref{example:conditional-independence}, and benchmark its performance against the  sequential method ECRT \citep{shaer2023model}. Note that we have tailored ECRT to fit our framework, that is,  both ECRT and DAVT-CIT employ the same network architecture for model fitting and use a single sample of $\tilde{U}$, at each step $t$ to estimate the betting scores $S_t$.

The type-I error and power of these two tests over 100 trials is plotted in~\Cref{fig:cit}. 
While both methods control type-I error at the required level $\alpha=0.05$, our test (DAVT-CIT) requires significantly fewer observations (on an average) to reject the null under $H_1$. The histograms of the two stopping times in~\Cref{fig:cit} further demonstrate the better sample-efficiency of DAVT-CIT.  

\subsection{Adversarial Attacks on ResNet50}
In this experiment, we evaluate the robustness of a ResNet50 model \citep{he2016deep}, refined on the CIFAR-10 dataset \citep{krizhevsky2009learning}, against adversarial attacks  as explained in  Example \ref{example:adversarial-attacks}. We employ the Fast Gradient Sign Method (FGSM) \citep{goodfellow2014explaining} for generating adversarial samples. Within this context, let $X$ represent the true CIFAR-10 images and $\Tau_{adv}(X)$ the biased ones produced with FGSM. Moreover, let the operator ${\Tau_h: \mathcal{X}\to \mathbb{R}^{10}}$ map an  image  to ResNet50 last layer~\footnote{Any intermediate layer of the network can be considered while defining an adversarial operator.}.  In this example, we consider ResNet50 to be robust against FGSM adversarial attacks if $\Tau_h(X)$ and $\Tau_h \circ \Tau_{adv}(X)$ have the same distribution. In other words, we will test: ${H_0: \Tau_h(X_t) \stackrel{d}{=} \Tau_h \circ \Tau_{adv}(X).}$

\begin{figure}
    \centering
    \includegraphics[width=0.78\columnwidth]{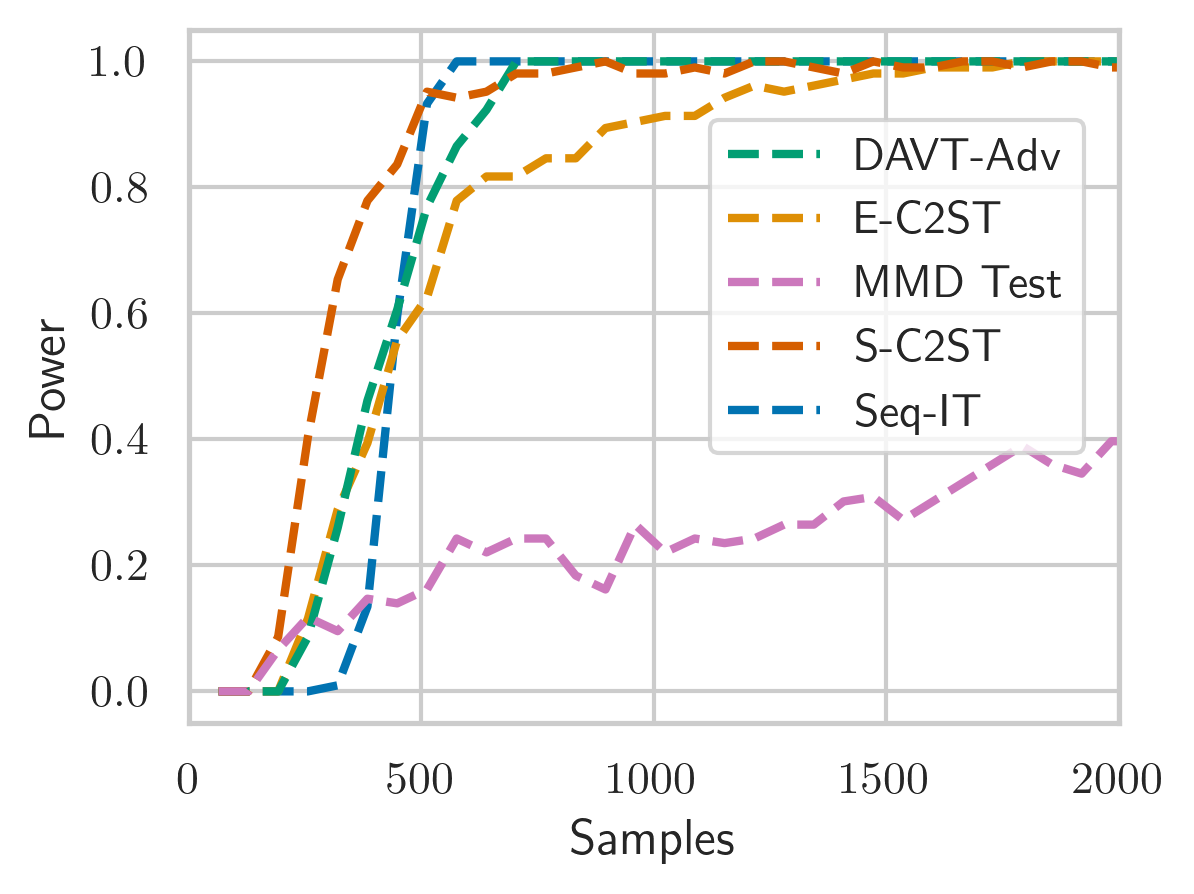}
    \caption{Power analysis for testing the adversarial robustness of ResNet50 trained on CIFAR-10. While S-C2ST initially outperforms in scenarios with limited data, DAVT-Adv (our method) and Seq-IT catch up and accelerate to reach maximum performance at a faster rate, leaving DAVT-Adv as the second best performing method.}
    \label{fig:cifar-power}
\end{figure}

We run 100 power experiments on batches of paired original and FGSM-altered images, with a sample size of 64. We compare DAVT-Adv with the two-sample test baselines: E-C2ST, Seq-IT, S-C2ST, and the MMD test.  \Cref{fig:cifar-power} shows that of the sequential methods, DAVT-Adv (ours) outperforms E-C2ST and Seq-IT until it reaches 60\% power. After that, Seq-IT achieves maximum power faster, leaving our method as the second best. Nevertheless, the non-sequential S-C2ST initially shows greater power than all other methods, but this advantage declines with increasing sample size.

\subsection{Rotation Invariance Testing} 
\begin{figure}
    \centering
    \includegraphics[width=\columnwidth]{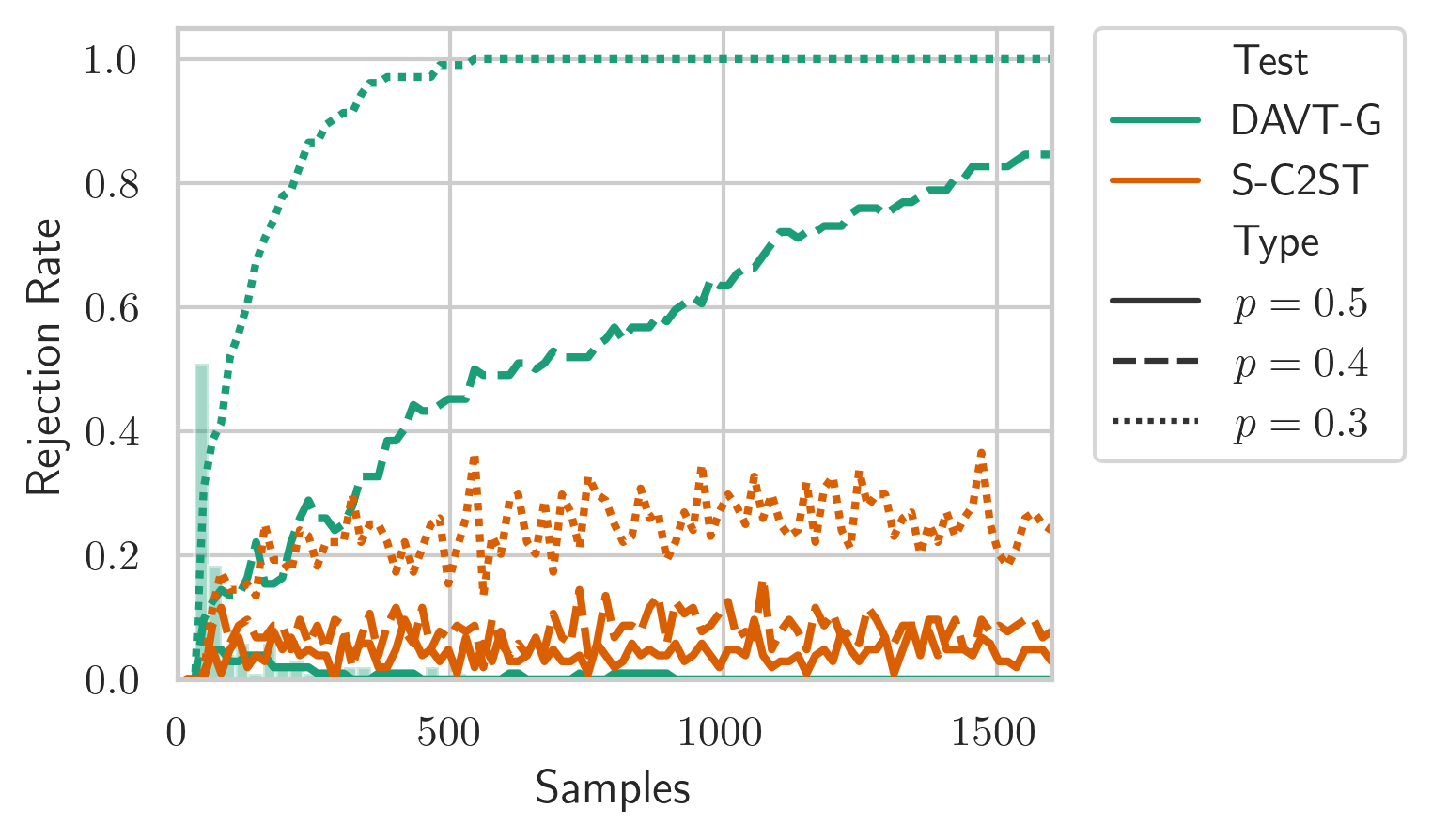}
    \caption{Power/Type I error analysis for the rotated 6 and 9 MNIST images for different mixing proportions $p=0.3, 0.4, 0.5.$ DAVT(ours) shows better power performance than the baseline S-C2ST combined with multiple testing corrections.}
    \label{fig:mnist-exp}
\end{figure}

We revisit Example \ref{example:group-invariance} and extend it the more general setting  described in \Cref{sec:rand-test}. More precisely, we consider that  at each time $t$ we are given batches of paired rotated images $Z=(X,Y)$ following the generative model $P_{X} = p\cdot P_6 + (1-p)\cdot P_9$ and $P_{Y} = p\cdot P_9 + (1-p)\cdot P_6$, where $p$ is the mixing weight. Here the distributions $P_6$ and $P_9$ represent the of the randomly rotated ``6'' and ``9" at angles in the set $\{90, 180, 270, 360\}$ degrees. This generative model is  apriori unknown to the practitioner who wants to check whether the two distributions, $P_{X}$ and $P_{Y}$,  remain equal under rotations of 90, 180, 270, or 360 degrees. Thus, by applying ideas from \Cref{sec:rand-test} we can form the null hypothesis:
\begin{align}
    H_0: \Tau_i(X)\stackrel{d}{=}Y\, \text{ for all } i \in \{90, 180, 270, 360\}
\end{align}
where the operators correspond to the specified rotations. Next, we adapt the test to our framework by introducing the two operator sets: ${\mathcal{O}_1= \{\Tau_{i}\circ \Tau_{\text{proj}}:\,i\in \{90, 180, 270, 360\} \}}$ and ${\mathcal{O}_2=\{ \Tau_{\text{proj,swap}} \}.}$ Thus, the above null hypothesis becomes equivalent to
\begin{align}
H_0: \Tau_i(Z)\stackrel{d}{=}\Tau_{\text{proj,swap}}(Z)\, \text{ for all } \Tau_i \in \mathcal{O}_1.
\end{align}
For constructing the two distributions, we use the MNIST dataset \citep{lecun2010mnist}. We run experiments for $p = 0.3, 0.4$ and $0.5$ in batches of 16 samples each.  \Cref{fig:mnist-exp} presents the power results for $p=0.3$ and $p=0.4$, and the type I error rate when $p=0.5$. As expected, the test successfully keeps the type I error under the significance threshold, and shows reduced power when considering the more challenging case of $p=0.4$, compared to $p=0.3$. 

We create a baseline test by applying S-C2ST to each hypothesis, each based on a single operator. This process involves training a different neural net for every hypothesis and then computing its associated $p$-value during testing. We then consolidate the four derived $p$-values using the Bonferroni correction.  \Cref{fig:mnist-exp} shows the rejection rates of this method for $p=0.3, 0.4, 0.5$ over time and highlights the lack of power of the method. This is a common effect when multiple correction procedures are used which vindicates the use of sequential tests.

\section{Discussion}
\label{sec:discussion}
    In this paper, we developed a unified deep learning based approach for constructing sequential tests for an abstract class of nonparametric testing problems. This class of problems includes various important applications ranging from two-sample testing, independence testing, to certifying adversarial robustness and group fairness of machine learning models.
    Our sequential test provides tight control over the type-I error, and is consistent under very mild conditions on the learning algorithm. 
    Through extensive empirical evaluation, we show that our general testing strategy, when instantiated to several practical applications, performs competitively with existing sequential tests specifically designed for those tasks. 

    Our work opens up several interesting directions for future work. For example, from a theoretical perspective, obtaining stronger guarantees on the performance of our test, perhaps by leveraging recent advances in deep learning theory~\citep{jacot2018neural}, is an important question. On the practical front, some interesting topics include  designing improved rules for computing the mini-batch betting scores (\Cref{app:implementation} in appendix), and identifying appropriate regularizations or modifications of the training objective functions, in order to learn better joint betting and payoff strategies.


\newpage 
\bibliographystyle{abbrvnat}
\bibliography{ref}

\begin{thebibliography}{27}
\providecommand{\natexlab}[1]{#1}
\providecommand{\url}[1]{\texttt{#1}}
\expandafter\ifx\csname urlstyle\endcsname\relax
  \providecommand{\doi}[1]{doi: #1}\else
  \providecommand{\doi}{doi: \begingroup \urlstyle{rm}\Url}\fi

\bibitem[Candes et~al.(2018)Candes, Fan, Janson, and Lv]{candes2018panning}
E.~Candes, Y.~Fan, L.~Janson, and J.~Lv.
\newblock {Panning for gold: ‘model-X’ knockoffs for high dimensional
  controlled variable selection}.
\newblock \emph{Journal of the Royal Statistical Society Series B: Statistical
  Methodology}, 80\penalty0 (3):\penalty0 551--577, 2018.

\bibitem[Chwialkowski et~al.(2015)Chwialkowski, Ramdas, Sejdinovic, and
  Gretton]{chwialkowski2015fast}
K.~P. Chwialkowski, A.~Ramdas, D.~Sejdinovic, and A.~Gretton.
\newblock Fast two-sample testing with analytic representations of probability
  measures.
\newblock \emph{Advances in Neural Information Processing Systems}, 28, 2015.

\bibitem[Cybenko(1989)]{cybenko1989approximation}
G.~Cybenko.
\newblock Approximation by superpositions of a sigmoidal function.
\newblock \emph{Mathematics of control, signals and systems}, 2\penalty0
  (4):\penalty0 303--314, 1989.

\bibitem[Darling and Robbins(1968)]{darling1968some}
D.~A. Darling and H.~Robbins.
\newblock Some nonparametric sequential tests with power one.
\newblock \emph{Proceedings of the National Academy of Sciences}, 61\penalty0
  (3):\penalty0 804--809, 1968.

\bibitem[Goodfellow et~al.(2014)Goodfellow, Shlens, and
  Szegedy]{goodfellow2014explaining}
I.~J. Goodfellow, J.~Shlens, and C.~Szegedy.
\newblock Explaining and harnessing adversarial examples.
\newblock \emph{arXiv preprint arXiv:1412.6572}, 2014.

\bibitem[Gretton et~al.(2012)Gretton, Borgwardt, Rasch, Sch{\"o}lkopf, and
  Smola]{gretton2012kernel}
A.~Gretton, K.~M. Borgwardt, M.~J. Rasch, B.~Sch{\"o}lkopf, and A.~Smola.
\newblock A kernel two-sample test.
\newblock \emph{The Journal of Machine Learning Research}, 13\penalty0
  (1):\penalty0 723--773, 2012.

\bibitem[Gr{\"u}nwald et~al.(2023)Gr{\"u}nwald, de~Heide, and
  Koolen]{grunwald2020safe}
P.~Gr{\"u}nwald, R.~de~Heide, and W.~M. Koolen.
\newblock Safe testing.
\newblock \emph{Journal of the Royal Statistical Society: Series B (to appear
  with discussion)}, 2023.

\bibitem[Hazan et~al.(2007)Hazan, Agarwal, and Kale]{hazan2007logarithmic}
E.~Hazan, A.~Agarwal, and S.~Kale.
\newblock Logarithmic regret algorithms for online convex optimization.
\newblock \emph{Machine Learning}, 69:\penalty0 169--192, 2007.

\bibitem[He et~al.(2016)He, Zhang, Ren, and Sun]{he2016deep}
K.~He, X.~Zhang, S.~Ren, and J.~Sun.
\newblock Deep residual learning for image recognition.
\newblock In \emph{Proceedings of the IEEE conference on computer vision and
  pattern recognition}, pages 770--778, 2016.

\bibitem[Hornik(1991)]{hornik1991approximation}
K.~Hornik.
\newblock Approximation capabilities of multilayer feedforward networks.
\newblock \emph{Neural networks}, 4\penalty0 (2):\penalty0 251--257, 1991.

\bibitem[Jacot et~al.(2018)Jacot, Gabriel, and Hongler]{jacot2018neural}
A.~Jacot, F.~Gabriel, and C.~Hongler.
\newblock Neural tangent kernel: Convergence and generalization in neural
  networks.
\newblock \emph{Advances in neural information processing systems}, 31, 2018.

\bibitem[Kim et~al.(2021)Kim, Ramdas, Singh, and
  Wasserman]{kim2021classification}
I.~Kim, A.~Ramdas, A.~Singh, and L.~Wasserman.
\newblock Classification accuracy as a proxy for two-sample testing.
\newblock \emph{The Annals of Statistics}, 49\penalty0 (1):\penalty0 411--434,
  2021.

\bibitem[Krizhevsky and Hinton(2009)]{krizhevsky2009learning}
A.~Krizhevsky and G.~Hinton.
\newblock Learning multiple layers of features from tiny images.
\newblock Master's thesis, Department of Computer Science, University of
  Toronto, 2009.

\bibitem[LeCun et~al.(2010)LeCun, Cortes, and Burges]{lecun2010mnist}
Y.~LeCun, C.~Cortes, and C.~Burges.
\newblock Mnist handwritten digit database.
\newblock \emph{ATT Labs [Online]. Available:
  http://yann.lecun.com/exdb/mnist}, 2, 2010.

\bibitem[Lehmann and Romano(2022)]{lehmann1986testing}
E.~L. Lehmann and J.~P. Romano.
\newblock \emph{Testing statistical hypotheses}.
\newblock Springer, 4th edition, 2022.

\bibitem[Lh{\'e}ritier and Cazals(2018)]{lheritier2018sequential}
A.~Lh{\'e}ritier and F.~Cazals.
\newblock A sequential non-parametric multivariate two-sample test.
\newblock \emph{IEEE Transactions on Information Theory}, 64\penalty0
  (5):\penalty0 3361--3370, 2018.

\bibitem[Lopez-Paz and Oquab(2017)]{lopezpaz2017revisiting}
D.~Lopez-Paz and M.~Oquab.
\newblock Revisiting classifier two-sample tests.
\newblock In \emph{International Conference on Learning Representations}, 2017.

\bibitem[Pandeva et~al.(2022)Pandeva, Bakker, Naesseth, and
  Forr{\'e}]{pandeva2022valuating}
T.~Pandeva, T.~Bakker, C.~A. Naesseth, and P.~Forr{\'e}.
\newblock E-valuating classifier two-sample tests.
\newblock \emph{arXiv preprint arXiv:2210.13027}, 2022.

\bibitem[Podkopaev and Ramdas(2023)]{podkopaev2023predictive}
A.~Podkopaev and A.~Ramdas.
\newblock Sequential predictive two-sample and independence testing.
\newblock \emph{Advances in neural information processing systems}, 2023.

\bibitem[Podkopaev et~al.(2023)Podkopaev, Bl{\"o}baum, Kasiviswanathan, and
  Ramdas]{podkopaev2023skit}
A.~Podkopaev, P.~Bl{\"o}baum, S.~Kasiviswanathan, and A.~Ramdas.
\newblock Sequential kernelized independence testing.
\newblock In \emph{International Conference on Machine Learning}, pages
  27957--27993. PMLR, 2023.

\bibitem[Ramdas et~al.(2022)Ramdas, Gr{\"u}nwald, Vovk, and
  Shafer]{ramdas2022game}
A.~Ramdas, P.~Gr{\"u}nwald, V.~Vovk, and G.~Shafer.
\newblock Game-theoretic statistics and safe anytime-valid inference.
\newblock \emph{arXiv preprint arXiv:2210.01948}, 2022.

\bibitem[Shaer et~al.(2023)Shaer, Maman, and Romano]{shaer2023model}
S.~Shaer, G.~Maman, and Y.~Romano.
\newblock Model-x sequential testing for conditional independence via testing
  by betting.
\newblock In \emph{International Conference on Artificial Intelligence and
  Statistics}, pages 2054--2086. PMLR, 2023.

\bibitem[Shafer(2021)]{shafer2021testing}
G.~Shafer.
\newblock Testing by betting: A strategy for statistical and scientific
  communication.
\newblock \emph{Journal of the Royal Statistical Society Series A: Statistics
  in Society}, 184\penalty0 (2):\penalty0 407--431, 2021.

\bibitem[Shah and Peters(2020)]{shah2020hardness}
R.~D. Shah and J.~Peters.
\newblock The hardness of conditional independence testing and the generalised
  covariance measure.
\newblock \emph{The Annals of Statistics}, 48\penalty0 (3):\penalty0
  1514--1538, 2020.

\bibitem[Shekhar and Ramdas(2023)]{shekhar2023nonparametric}
S.~Shekhar and A.~Ramdas.
\newblock Nonparametric two-sample testing by betting.
\newblock \emph{IEEE Transactions on Information Theory}, 2023.

\bibitem[Szegedy et~al.(2014)Szegedy, Zaremba, Sutskever, Bruna, Erhan,
  Goodfellow, and Fergus]{szegedy2014intriguing}
C.~Szegedy, W.~Zaremba, I.~Sutskever, J.~Bruna, D.~Erhan, I.~Goodfellow, and
  R.~Fergus.
\newblock Intriguing properties of neural networks.
\newblock In \emph{2nd International Conference on Learning Representations,
  ICLR 2014}, 2014.

\bibitem[Ville(1939)]{ville1939etude}
J.~Ville.
\newblock Etude critique de la notion de collectif.
\newblock \emph{Gauthier-Villars, Paris.}, 1939.

\end{thebibliography}

\newpage 

%

%

\onecolumn
\aistatstitle{Deep anytime-valid hypothesis testing: \\
Supplementary Materials}
   
    \section{More examples}
    \label{app:more-examples} 

        \begin{example}[Independence Testing]
        \label{example:independence-testing}
            Independence testing is another well-studied problem in statistics, where given observations $\{(X_i, Y_i): 1 \leq i \leq n\}$ drawn \iid from a distribution $P_{XY}$ on a product space $\mc{X} \times \mc{Y}$, we want to test whether $P_{XY} = P_X \times P_Y$ or not. By working with two pairs of observations at a time, we can again describe the null as being invariant to an operator. In particular, let given $Z_1 = (X_1, Y_1)$ and $Z_2 = (X_2, Y_2)$, let $\Tau$ denote the operator that maps $(Z_1, Z_2)$ to $(Z_1', Z_2')$, with $Z_1' = (X_1, Y_2)$ and $Z_2' = (X_2, Y_1)$. Clearly, the distribution of $(Z_1, Z_2)$ is the same as that of $(Z_1', Z_2')$ under the null, while this invariance to $\Tau$ is broken under the alternative. 
        \end{example}
    
        \begin{example}[Symmetry testing]
            \label{example:symmetry-testing}
            In the simplest version of this problem, we consider real-valued observations~(that is, $\mc{Z}=\reals$), and the operators $\Tau_2 = \identity$, and $\Tau_1 = \flip$, where the operator $\flip$  simply flips the observations about the origin; that is, $\flip(x) = -x$. The resulting null hypothesis  asserts that $P_Z$ is symmetric about the origin. 
            The same formulation also covers other kinds of symmetry, such as rotational invariance, or invariance to horizontal or vertical flips in the case of images.        
        \end{example}

        \begin{example}[Group Fairness]  
            \label{example:group-fairness}     Group fairness, sometimes referred to as demographic or statistical fairness, is a research area in machine learning that focuses on how machine learning models perform across different demographic groups. The main goal is to ensure that a model's performance is consistent across predefined groups, avoiding situations where the model may disproportionately benefit or harm a particular group. 
             
             A typical application in this context would be testing whether an ML model is racially biased. For example, suppose a trained ML recommendation model $h$ is used to predict which candidates are most likely to succeed in a job. The company running this model wants to ensure that it is not racially biased. To achieve this, they categorize applicants into $p$ ethnic groups and then evaluate whether the model $h$ produces consistent results across all $p$ groups. Let $Y$ be the categorical random variable indicating the demographic group and $X$ be a random vector collecting the rest of the applicant's covariates.  The associated statistical test is 
            \begin{align}
                H_0: h(X) \perp \!\!\! \perp Y
            \end{align}
  Thus, by using  \Cref{example:independence-testing}, we can design a test  with respect to the defined null based on our framework.
        \end{example}
\begin{example}[Group invariance testing]
\label{example:rotation-invariance}
Suppose we have a collection of images $Z$ containing equilateral triangles. Each edge of these triangles is colored either blue or green.  A practitioner would like to find out if the edges of the triangles are colored without any particular pattern, or if some hidden rule controls their coloring. To do this, we will examine whether the triangles remain the same when rotated 120 or 240 degrees.  We will therefore introduce a set of operators that represent the aforementioned rotations: $\Tau_{120}$ and $\Tau_{240}$. Next, we formulate the following \textit{composite} null hypothesis:
\begin{align}
    H_0: \Tau_{120}(Z)=Z  \text{ and }  \Tau_{240}(Z)=Z 
\end{align}
Thus, we want to test whether the distribution of $Z$ remains invariant with respect to the two operators $\Tau_{120}$ and $\Tau_{240}$. 
\end{example}

    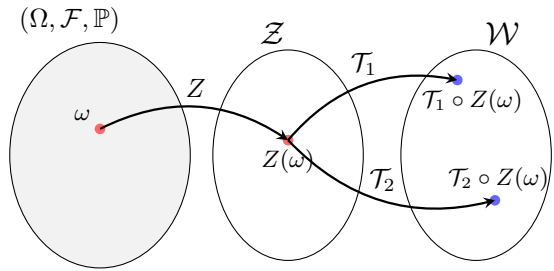
\begin{figure}[hbt!]
        \centering
        \def\figwidth{0.5\columnwidth}
        \def\figheight{0.25\columnwidth} 
         \begin{tikzpicture}[>=stealth, scale=0.5]
              \def\xRadius{2.4}
              \def\yRadius{3}
              \draw[fill=gray!10] (-4, 0) ellipse (\xRadius cm and \yRadius cm);
              \node[anchor=south west] at (-4-\xRadius, \yRadius) {$(\Omega, \mathcal{F}, \mathbb{P})$};
              \draw (1, 0) ellipse (2 cm and 2.8 cm);
              \node[anchor=south west] at (0,2.7) {\large $\mathcal{Z}$};
              \draw (6, 0) ellipse (2 cm and 2.8 cm);
              \node[anchor=south west] at (6,2.7) {\large $\mathcal{W}$};

            \fill[color=red!60] (-4, 0.7) circle (4pt); 
            \node[above left] at (-4, 0.7) {\small$\omega$};

            \fill[color=red!60] (1, 0.4) circle (4pt); 
            \node[below] at (1, 0.4) {\small$Z(\omega)$};
            \fill[color=blue!60] (5.5, 2.0) circle (4pt); 
            \fill[color=blue!60] (6.5, -1.2) circle (4pt); 
            \node[below] at (5.9, 2.0) {\small $\Tau_1 \circ Z(\omega)$};
            \node[above] at (6.6, -1.2) {\small $\Tau_2 \circ Z(\omega)$};
            
            \draw[->, bend left=30, line width = 0.85pt] (-4, 0.7) to node[above] {$Z$} (1, 0.4);
            \draw[->, bend left=30, line width = 0.85pt] (1, 0.4) to node[above] {$\Tau_1$} (5.5, 2.0) ;
            \draw[->, bend right=30, line width = 0.85pt] (1, 0.4) to node[above] {$\Tau_2$} (6.5, -1.2) ;
        \end{tikzpicture}
        \caption{Let $Z$ denote a $\mc{Z}$-valued random variable on an underlying probability space $(\Omega, \mc{F}, \mathbb{P})$.  By definition, the distribution of the $Z$ is equal to $P_Z = \mathbb{P} \circ Z^{-1}$. The two black curved lines from $\mc{Z}$ to $\mc{W}$ denote the operators $\Tau_i$ for $i \in \{1, 2\}$, used to characterize the class of null distributions. In particular, the distribution of the resulting $\mc{W}$-valued random variables are $\mathbb{P} \circ (\Tau_i \circ Z)^{-1} = \mathbb{P} \circ Z^{-1} \circ \Tau_i^{-1} = P_Z \circ \Tau_i^{-1}$, and the  null hypothesis of our abstract testing problem states that the two distributions, $P_Z \circ \Tau_1^{-1}$ and $P_Z \circ \Tau_2^{-1}$, are the same.}
        \label{fig:problem-statement}
    \end{figure}

    \section{Batch test based on sample-splitting}
    \label{sec:batch-test-sample-splitting}
        We can also construct a batch test (also called a fixed sample-size test) for the abstract testing problem~\eqref{eq:abstract-problem-def} using the idea of sample splitting. In particular, let $\mc{D} = \{Z_i: 1 \leq i \leq n\}$ denote the set of observations, which are then split into two equal halves, $\mc{D}_1 = \{Z_1, \ldots, Z_{n/2}\}$ and $\mc{D}_2 = \{Z_{n/2+1}, \ldots, Z_n\}$. We use the first split, $\mc{D}_1$, to train an ML model~(usually a DNN) with the objective of maximizing the growth rate: 
        \begin{align}
            \thetahat \in \argmin_{ \theta \in \Theta} \; -\frac{2}{n} \sum_{Z_i \in \mc{D}_1} \log \lp 1 +  \gtilde_{\theta}(Z_i)  \rp. \label{eq:theta-hat}
        \end{align}
        Next, we use the learned parameter on the second split to construct the test statistic 
        \begin{align}
        \label{eq:test-statistic}
            E_n =  \prod_{Z_i \in \mc{D}_2} \lp  1 + \gtilde_{\thetahat} (Z_i) \rp. \label{eq:batch-e-value}
        \end{align}
        We expect this statistic $E_n$ to be small under the null, and thus we can use it to define a test for~\eqref{eq:abstract-problem-def} that rejects the null for large values of $E_n$. Our next result analyzes the performance of such a test.  
        \begin{proposition}
            \label{prop:batch-test} Suppose the learning algorithm  ensures that 
            \begin{align}
                \liminf_{n \to \infty} \frac{\mathbb{E}[\log(1+\gtilde_{\thetahat}(Z))|\mc{D}_1]}{4 c\sqrt{\log n /n }} \stackrel{a.s.}{>} 1, \quad \text{under } H_1, \label{eq:assump-batch-test}
            \end{align}
            where $c$ is a universal constant. 
            Then, the test $\Psi(Z^n) = \ind_{E_n \geq 1/\alpha}$, that rejects the null if $E_n$ exceeds $1/\alpha$, satisfies the following properties: 
            \begin{align}
                \mathbb{E}_{H_0}\lb \Psi(Z^n) \rb \leq \alpha, \quad \text{and} \quad \lim_{n \to \infty} \mathbb{E}_{H_1}\lb \Psi(Z^n) \rb = 1. 
            \end{align}
            That is, $\Psi$ is a consistent, level-$\alpha$ test for~\eqref{eq:abstract-problem-def}.
       \end{proposition}
       The proof of this result is in~\Cref{proof:batch-test} of the appendix. 
   
        Note that the test statistic $E_n$ is an \emph{e-variable}~\citep{grunwald2020safe}; which is a nonnegative random variable with an expected value no larger than $1$ under the null. As a result, the test $\Psi$ is valid under \emph{optional continuation}. That is, suppose we compute $E_n$ using a dataset $\mc{D}$, and its value turns out to be smaller than $1/\alpha$. Since the test $\Psi$ based on $\mc{D}$ is inconclusive, we may decide to collect $m$ further observations, $\mc{D}'$, and use it to compute $E_m'$. We can combine the evidence from the two experiments easily by simply rejecting the null if $E_n \times E_m'$ exceeds $1/\alpha$, without violating type-I error guarantees. This is a simple consequence of the fact that $\mathbb{E}[E_n E_m'] = \mathbb{E}[E_n \mathbb{E}[E_m'|\mc{D}]] \leq \mathbb{E}[E_n] \leq 1$. 

\section{Deferred Proofs}
\label{app:proofs}
In this section, we present the proofs of the two results analyzing the performance of our batch-test $\Psi$~(\Cref{prop:batch-test}), and the sequential test~$\gamma$~(\Cref{prop:sequential-test}). 
    \subsection{Proof of Proposition~\ref{prop:batch-test}} 
    \label{proof:batch-test}
        \paragraph{Type-I error control.} The proof of the type-I error control follows from an application of Markov's inequality. More specifically, the expected value of $E_n$ can be expressed as follows: 
        \begin{align}
            \mathbb{E}[E_n]  &= \mathbb{E}[\mathbb{E}[E_n|\mc{D}_1]] = \mathbb{E}\lb \mathbb{E} \lb \prod_{Z_i \in \mc{D}_2}\lp 1 + \gtilde_{\thetahat}(Z_i) \rp \mid \mc{D}_1\rb \rb \\
            & = \mathbb{E}\lb  \prod_{Z_i \in \mc{D}_2}\lp 1 + \mathbb{E} \lb \gtilde_{\thetahat}(Z_i) \mid \mc{D}_1\rb\rp  \rb,  \label{eq:proof-batch-1}
        \end{align}
        where $\mc{D}_1$ and $\mc{D}_2$ denote the two splits of the dataset $\mc{D}$. Recall that the parameter $\thetahat$ trained on the first split $\mc{D}_1$, and thus $\gtilde_{\thetahat}$ can be treated as a constant function in the conditional expectation above. The equality in~\eqref{eq:proof-batch-1} uses the fact that dataset $\mc{D}$, and hence the split $\mc{D}_2$ consists of \iid data-points. Finally, we have 
        \begin{align}
            \mathbb{E} \lb \gtilde_{\thetahat}(Z_i) \mid \mc{D}_1\rb =0, 
        \end{align}
        under $H_0$ for any $Z_i \in \mc{D}_2$, which implies that 
        \begin{align}
            \mathbb{E}[E_n] = 1, \quad \text{under } H_0. 
        \end{align}
        This immediately implies the type-I error guarantee of our test $\Psi$, since 
        \begin{align}
            \mathbb{P}_{H_0} \lp \Psi(Z^n) = 1 \rp  = \mathbb{P}_{H_0} \lp E_n \geq 1/\alpha \rp \leq  \frac{\mathbb{E}_{H_0}[E_n]}{1/\alpha} = \frac{1}{1/\alpha} = \alpha. 
        \end{align}
        The inequality above is due to Markov's inequality. 
        
        \paragraph{Consistency.} For proving the consistency of our test, we need some additional notation: 
        \begin{align}
            v_i = \log \lp 1 + \gtilde_{\thetahat}(Z_{n/2 + i}) \rp, \; \text{for } i \in \{1, \ldots, n/2\}, \quad \text{and} \quad A_n \defined \mathbb{E}\lb \gtilde_{\thetahat}(Z) | \mc{D}_1 \rb, 
        \end{align}
        where $Z \sim P_Z$ is independent of $\mc{D}_1$.  Now, observe the following 
        \begin{align}
            \mathbb{P}\lp \Psi(Z^n) = 1 \rp & = \mathbb{P}\lp \frac{2\log E_n}{n}  \geq \frac{2\log(1/\alpha)}{n} \rp  = \mathbb{P}\lp \frac{2}{n} \sum_{i=1}^{n/2} v_i \geq \frac{2 \log(1/\alpha)}{n} \rp \\
            &= \mathbb{P}\lp  A_n +  \frac{2}{n} \sum_{i=1}^{n/2} (v_i - A_n)\geq \frac{2 \log(1/\alpha)}{n} \rp. \label{eq:batch-test-proof-1}
        \end{align}
        Now, we introduce the event $G_n = \{ |(2/n) \sum_{i=1}^{n/2} v_i - A_n| \leq c\sqrt{4 \log n /n } \}$, for $c = 2\log(1/(1-2q))$ and $q \in (0, 1/2)$ is the upper bound on $|g_\theta(x)|$ for all $x, \theta$ assumed in~\Cref{sec:proposed-approach}. Note that for each $i$, the random variable $v_i - A_n$ is bounded in $[-c/2,c/2]$, with mean $0$. This means that, by an application of Hoeffding's inequality, we get 
        \begin{align}
            \mathbb{P}\lp G_n^c \rp \leq \frac{2}{n^2}, \quad \text{which implies that} \quad \sum_{n=2}^{\infty} \mathbb{P}(G_n^c) < \infty. \label{eq:batch-test-proof-2}
        \end{align}
        Returning to~\eqref{eq:batch-test-proof-1}, we now get 
        \begin{align}
            \mathbb{P}\lp \Psi(Z^n) = 1 \rp &\geq  \mathbb{P}\lp  \lbr A_n +  \frac{2}{n} \sum_{i=1}^{n/2} (v_i - A_n)\geq \frac{2 \log(1/\alpha)}{n} \rbr \cap G_n \rp \\ 
            &\geq  \mathbb{P}\lp  \lbr A_n \geq \frac{2 \log(1/\alpha)}{n}  + 2c \sqrt{\frac{\log n}{n} } \rbr \cap G_n \rp. 
        \end{align}
        The second inequality above uses the fact that under the event $G_n$, the term $(2/n) \lp  \sum_{i=1}^{n/2} v_i - A_n \rp$ is lower bounded by $- 2c \sqrt{\log n/n}$. 
        
        For large enough values of $n$, the term $2 \log(1/\alpha)/n$ is smaller than $2c\sqrt{\log n /n }$. Using this fact, we obtain 
        \begin{align}
            \mathbb{P}(\Psi(Z^n) = 1) & \geq \mathbb{P}\lp H_n \cap G_n \rp = \mathbb{E}[\ind_{H_n}\,\ind_{G_n}], \quad \text{with } H_n \defined \lbr B_ n \geq 4c \sqrt{ \frac{\log n }{n}} \rbr. \label{eq:batch-test-proof-3}
        \end{align}
        Now, taking the limiting value of the probability of detection, we get
        \begin{align}
            1 \geq \liminf_{n \to \infty} \mathbb{P}\lp \Psi(Z^n) =1 \rp \geq \liminf_{n \to \infty} \mathbb{E}[\ind_{H_n}\,\ind_{G_n}]  \geq \mathbb{E}[\liminf_{n \to \infty} \ind_{H_n}\,\ind_{G_n}], 
        \end{align}
        where the last inequality follows by an application of Fatou's Lemma. 

        To complete the proof, it suffices to show that $\ind_{H_n} \ind_{G_n} \convas 1$, which would imply that $\lim_{n \to \infty} \mathbb{P}\lp \Psi(Z^n) = 1 \rp = 1$. We show this in two steps: 
        \begin{itemize}
            \item From~\eqref{eq:batch-test-proof-2}, and an application of (the first) Borel-Cantelli Lemma, we know that 
            \begin{align}
                \mathbb{P}\lp G_n^c \text{infinitely often} \rp = \mathbb{P}\lp \cap_{n = 1}^{\infty} \cup_{m \geq n} G_m^c \rp = 0. 
            \end{align}
            On taking the complement of the event above, we get 
            \begin{align}
                \mathbb{P}\lp \cup_{n \geq 1} \cap_{m \geq n} G_n \rp = 1, \quad \text{which implies that} \quad \ind_{G_n} \convas 1.   \label{eq:batch-test-proof-4}
            \end{align}
            \item For the final step, we use the assumption about the learning algorithm made in~\Cref{prop:batch-test}. In particular,  the assumption  that 
            \begin{align}
                \liminf_{n \to \infty} \frac{A_n}{4c \sqrt{\log n/n}} \geqas 1 \quad \text{implies} \quad 
                \ind_{H_n} \convas 1. \label{eq:batch-test-proof-5}
            \end{align}
        \end{itemize}
        Together,~\eqref{eq:batch-test-proof-3} and~\eqref{eq:batch-test-proof-4} imply the required condition that $\ind_{G_n} \ind_{H_n} \convas 1$. This completes the proof.

    \subsection{Proof of~Proposition~\ref{prop:sequential-test}} 
    \label{proof:sequential-test}
        \paragraph{Type-I error control.} The type-I error control is a consequence of the fact that the process $\{W_t: t \geq 1\}$ is a non-negative martingale with an initial value of $1$. As a result, we have 
        \begin{align}
            \mathbb{P}(\gamma < \infty) = \mathbb{P}\lp \exists t \geq 1: W_t \geq 1/\alpha \rp \leq \frac{\mathbb{E}[W_0]}{1/\alpha} = \alpha, 
        \end{align}
        due to an application of Ville's inequality~\citep{ville1939etude}. 

        \paragraph{Consistency.} Recall that we use $t$ to denote the mini-batch counter, and $b$ to denote the size of each mini-batch. To prove the consistency of this test, we begin by observing that 
        \begin{align}
            \mathbb{P}(\gamma = \infty) = \mathbb{P}\lp \cap_{t \geq 1} \{ \gamma > t \} \rp \leq \mathbb{P}\lp \gamma > t \rp,  
        \end{align}
        for any arbitrary $t$. Taking the limit, this implies that 
        \begin{align}
            \mathbb{P}\lp \gamma = \infty \rp \leq \limsup_{t \to \infty} \mathbb{P}\lp \gamma > t \rp. 
        \end{align}
        To complete the proof, we will show that the RHS above is equal to $0$. As in the proof of~\Cref{prop:batch-test}, we introduce the notation 
        \begin{align}
            v_i = \sum_{Z \in B_i} \log\lp 1 + \gtilde_{\theta_{i-1}}(Z) \rp, \quad \text{and} \quad A_i = \mathbb{E}\lb v_i | \mc{F}_{i-1} \rb = b \times \mathbb{E}[\log(1 + \gtilde_{\theta_{i-1}}(Z))|\mc{F}_{i-1}], \label{eq:seq-test-proof-10}
        \end{align}
        where  $\mc{F}_{i-1}  = \sigma \lp \cup_{j=1}^{i-1} B_j \rp$ is the $\sigma$-algebra generated by the first $i-1$ batches of observations. 
        Then, we have 
        \begin{align}
            \mathbb{P}\lp \gamma > t \rp \leq \mathbb{P} \lp \frac{\log W_t}{t} < \frac{\log (1/\alpha)}{t} \rp  = \mathbb{P} \lp \frac{1}{t} \sum_{i=1}^t v_i - A_i + \frac{1}{t} \sum_{i=1}^t A_i < \frac{\log(1/\alpha)}{t} \rp.  \label{eq:seq-test-proof-1}
        \end{align}
        Now, observe that the process $\{v_i - A_i: i \geq 1\}$ is a bounded martingale difference sequence. Hence, an application of Azuma's inequality gives us 
        \begin{align}
            \mathbb{P}\lp G_t^c \rp \leq \frac{2}{t^2}, \quad \text{with} \; G_t \defined \lbr \lv \frac{1}{t} \sum_{i=1}^t v_i - A_i \rv \leq cb\sqrt{\frac{\log t}{t}} \rbr, 
        \end{align}
        where we have $c= 2 \log(1/(1-2q))$, and $q \in (0, 1/2)$ is the upper bound on $|g_\theta(x)|$ for all $x, \theta$ assumed in~\Cref{sec:proposed-approach}. 
        Combining the above result with~\eqref{eq:seq-test-proof-1},   we get 
        \begin{align}
            \mathbb{P}\lp \gamma > t \rp & \leq  \mathbb{P} \lp \lbr \frac{1}{t} \sum_{i=1}^t A_i < \frac{\log(1/\alpha)}{t} + \lv \frac{1}{t} \sum_{i=1}^t v_i - A_i \rv \rbr \cap G_t\rp + \mathbb{P}\lp G_t^c  \rp  \\ 
            & \leq  \mathbb{P} \lp \lbr \frac{1}{t} \sum_{i=1}^t A_i < \frac{\log(1/\alpha)}{t} + cb\sqrt{\frac{\log t}{t}} \rbr \cap G_t\rp + \mathbb{P}\lp G_t^c  \rp \label{eq:seq-test-proof-2} \\
            & \leq  \mathbb{P} \lp \frac{1}{t} \sum_{i=1}^t A_i <  2cb\sqrt{\frac{\log t}{t}}\rp + \frac{2}{t^2} \label{eq:seq-test-proof-3}. 
        \end{align}
        In the last inequality, we used the fact that for sufficiently large $t$, the term $\log(1/\alpha)/t$ is smaller than $c\sqrt{\log t/t}$, and that $\mathbb{P}(G_t^c) \leq 2/t^2$. By taking the limit in~\eqref{eq:seq-test-proof-3}, we obtain 
        \begin{align}
            \mathbb{P}(\gamma = \infty) \leq \limsup_{t \to \infty} \mathbb{P}\lp \gamma > t \rp \leq \limsup_{t \to \infty} \mathbb{E}\lb \ind_{H_t} \rb, \quad \text{where} \quad H_t \defined \lbr \frac{1}{t} \sum_{i=1}^t A_i < 2cb \sqrt{\frac{\log t}{t}} \rbr. \label{eq:seq-test-proof-4}
        \end{align}
        From the properties of Cesaro means, we know that
        \begin{align}
            \liminf_{n \to \infty} \frac{1}{n} \sum_{i=1}^n A_i \stackrel{a.s.}{\geq} \liminf_{n \to \infty} A_n, 
        \end{align}
        which implies 
        \begin{align}
            \liminf_{t \to \infty}  \frac{\frac{1}{t} \sum_{i=1}^t A_i}{2cb\sqrt{\log t/t}} \stackrel{a.s.}{\geq} \liminf_{t \to \infty} \frac{(A_t/b)}{2c\sqrt{\log t/t}} \geqas 1. \label{eq:seq-test-proof-5}
        \end{align}
        The last (strict) inequality is due to the assumption on the learning algorithm and noting that $\lim_{t \to \infty} \lp \sqrt{\log t/t}\rp/\lp \sqrt{\log (t-1)/(t-1)} \rp = 1$. This condition implies that $\ind_{H_n} \convas 0$, which by the Bounded convergence theorem~(or the continuity of probability) leads to 
        \begin{align}
            \mathbb{P}(\tau = \infty) \leq \limsup_{t \to \infty} \mathbb{E}[\ind_{H_t}] = 0, 
        \end{align}
        under the alternative. Hence we have proved the required statement that $\mathbb{P}(\tau < \infty) = 1$ under the alternative. 
        
\section{Experimental Details}
\label{app:implementation}
We implemented our and the baseline models in pytorch, and the chosen optimizer was Adam.  All calculations were performed on a local computing cluster offering 8 TitanX GPU nodes. The average time for each run is 6 min. Thus, the total compute time for the DNN experiments in the main paper is approximately $6\cdot 33\cdot 100$ minutes or 330 GPU hours. 

In our experiments, we draw mini-batches with consistent sample sizes and then apply \Cref{algo:sequential-test}. We trained all models with early stopping coupled with a low patience threshold as a safeguard against potential model overfitting.   In this section, we will explain the implementation and training procedure in detail for each experiment: two-sample testing  in Section \ref{app:blob}, adversarial robustness in Section \ref{app:cifar}, group invariance in Section \ref{app:mnist}, CIT in Section \ref{app:cit}. The hyperparameters are generally selected so the resulting tests have small stopping times. The search space for the learning rate is: $\{0.0005,0.0001, 0.005, 0.001\}$ and for the patience: $\{2,5,10\}.$
\subsection{Implementation of  DAVT}
    
    The training procedure with early stopping always uses the last batch as a validation set. Note that at the beginning of the procedure, the first batches $B_1$ and $B_2$ are used for training and validation. We also implement a variation of the loss function that is used for optimization, i.e.
    \begin{align}
        \theta_t \in\argmax_{\theta\in\Theta}\sum_{l=1}^t\sum_{Z\in B_l} \log\left( 1+\sigma\left(g_{\theta}\circ\Tau_1(Z)-g_{\theta}\circ\Tau_2(Z)\right)\right),
    \end{align}
    where $\sigma:\mathbb{R}\to[-1,1]$ is an monotone increasing function with $\sigma(-x) = -\sigma(x)$. 
\subsection{Baselines}
\label{app:baselines}
Here, we provide an overview of the implemented baseline methods.
\begin{itemize}
    \item E-C2ST \citep{lheritier2018sequential,pandeva2022valuating} is a sequential two-sample test based on the M-split likelihood ratio testing where the trained DNN maximizes the data log-likelihood  of the previous batches under the alternative.
    \item Seq-IT \citep{podkopaev2023predictive} extends E-C2ST by formalizing the test in the "testing by betting" framework and thus coupling the payoff to a betting strategy such as ONS. The ONS betting strategy $\lambda_t$ is computed samplewise in the original paper. To make Seq-IT comparable to our framework, we update the betting strategy batch-wise, resulting in a constant $\lambda_b$ for the entire batch.
    \item S-C2ST \citep{lopezpaz2017revisiting,kim2021classification} is a non-sequential classifier two-sample test that uses  permutation testing  for constructing the p-value. 
    \item MMD Test \citep{gretton2012kernel} is a kernel-based test utilizing permutation testing. To adapt MMD for testing image data we can use the capabilities of a trained neural network to act as a feature extractor. We then use the extracted features to perform a kernel MMD test.
    \item ECRT \citep{shaer2023model} is the only conditional independent test on the list. It is also based on the ``testing by betting" paradigm. The difference between DAVT-CIT and ECRT is the presence of a betting strategy linked to the universal portfolio optimization paradigm. 
\end{itemize}
\subsection{Two-Sample Testing}
\label{app:blob} 
\begin{figure}[!h]
    \centering
    \includegraphics[scale=0.3]{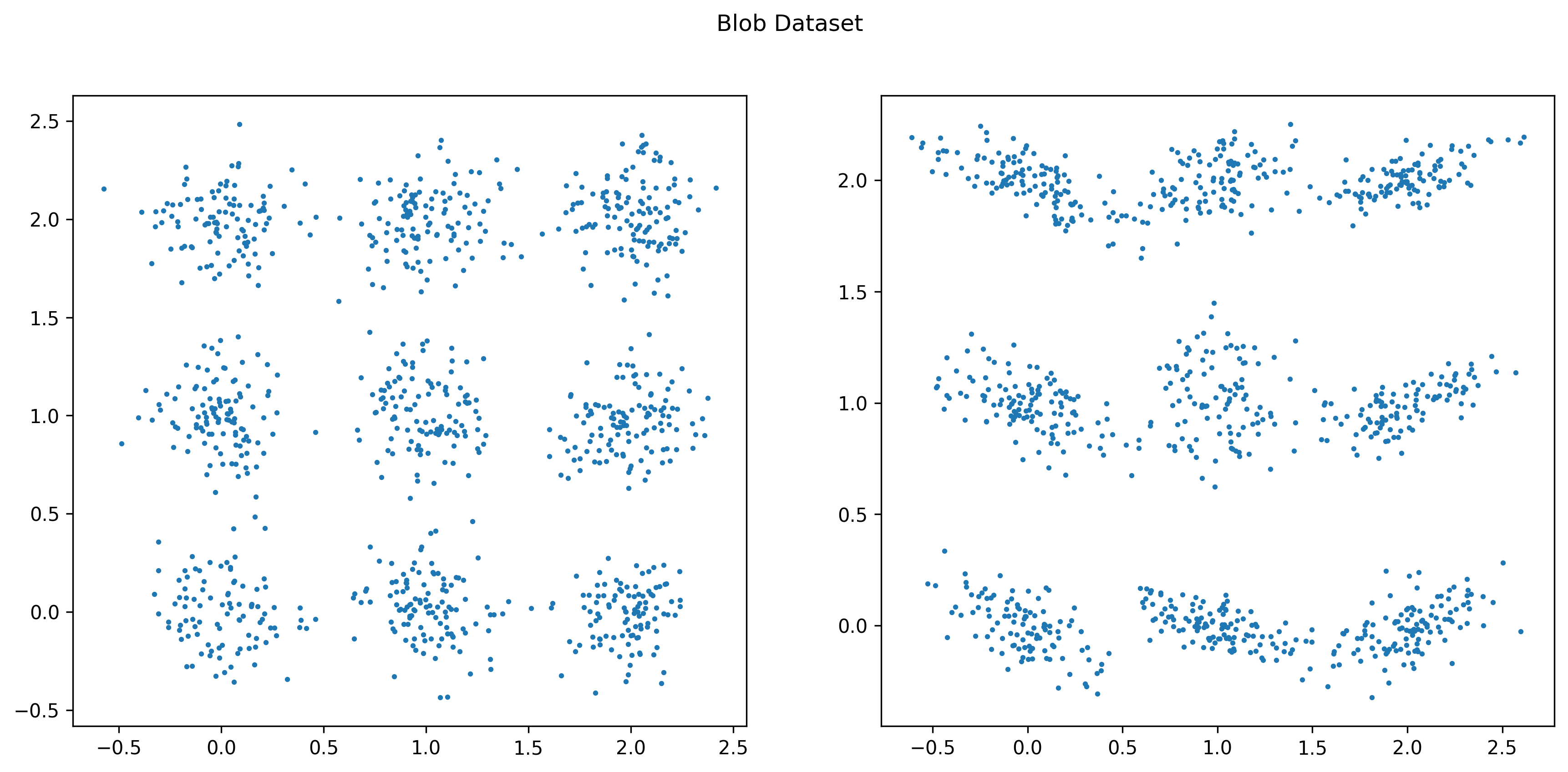}
    \caption{Blob Dataset}
    \label{fig:blob}
\end{figure}

The Blob dataset is a synthetic benchmark dataset for two sample tests. It is a challenging dataset due to the overlapping modes of the two classes. Figure~\ref{fig:blob} provides a visualization of the class distributions.

All trained DNN models follow the network architecture in Table~\ref{tab:blob}. The net contains three linear layers alternating with a LayerNorm and ReLU activation function. All models are trained for a maximum of 500 with early stopping with respect to the loss on the validation set with patience ten and a learning rate of 0.0005. 

\subsubsection{Lower Capacity DNNs}
\label{subsub:low}
\begin{figure}
    \centering
    \includegraphics[width=\textwidth]{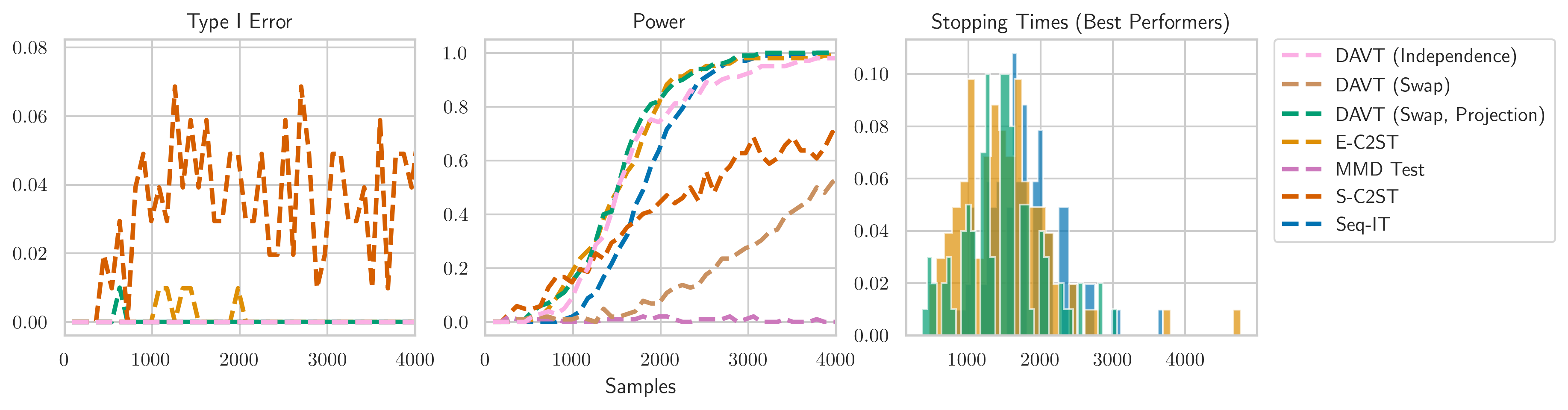}
    \caption{Blob Experiments with lower capacity DNN.}
    \label{fig:blob-low}
\end{figure}

 We also performed additional experiments where the chosen DNN has a lower capacity, i.e. the size of the hidden layers is 10 instead of 30 as in Table \ref{tab:blob}. The results presented in Figure \ref{fig:blob-low} show that the tests' tendencies and ranking in terms of their performance are maintained. Interestingly, the independence test performs better here than in the previous experiments. However, all methods need twice as many samples to achieve maximum power. This illustrates the role of the DNN in the performance of all tests. 
\subsubsection{Two-sample Test for Dependent $X$ and $Y$ }
\label{subsub:indep}
\begin{figure}
    \centering
    \includegraphics[width=0.39\columnwidth]{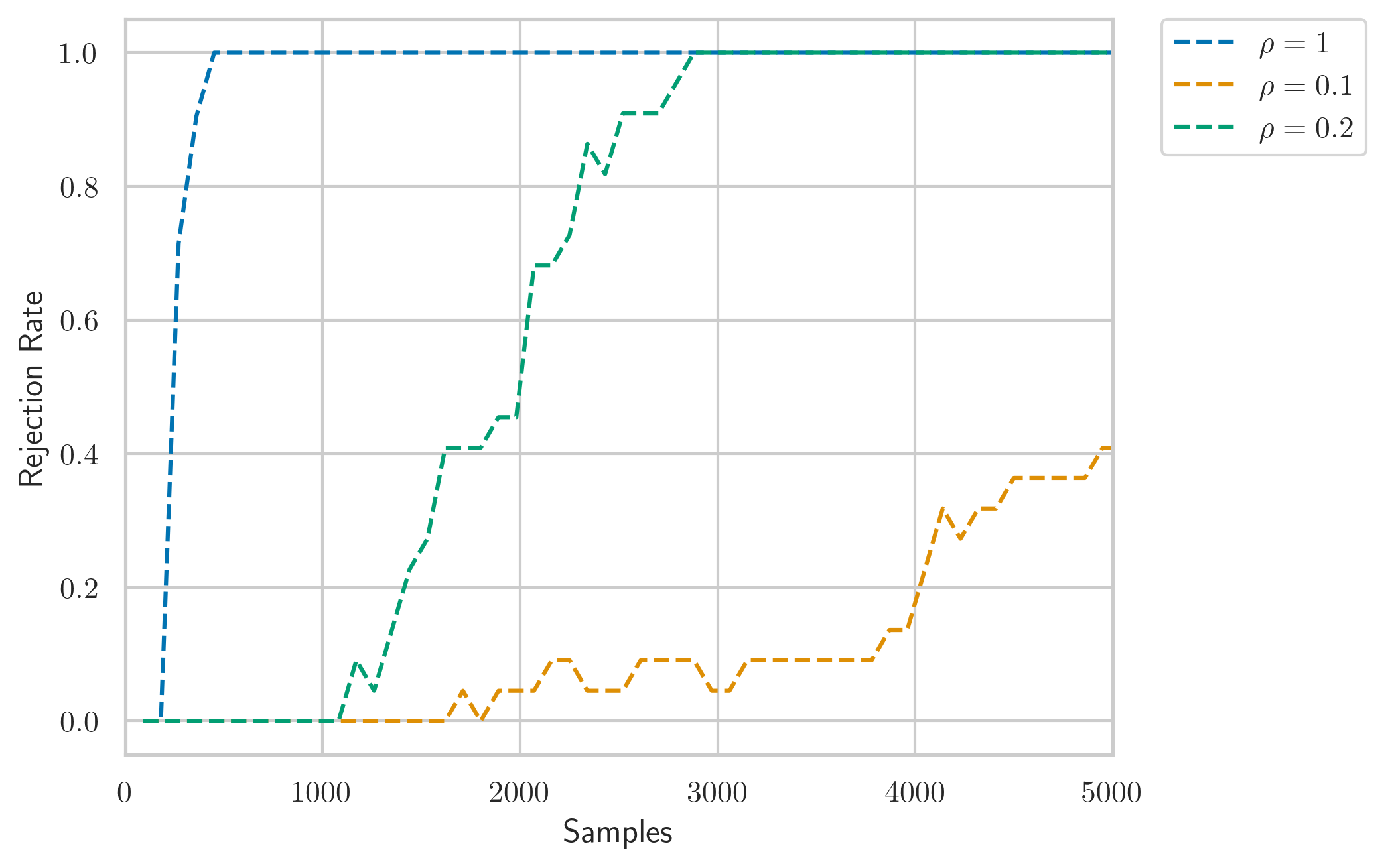}
    \caption{Rejection rate for the Blob two-sample test with swap operator. Here, $\rho$ is the correlation between the first dimensions of $X$ and $Y$. The stronger the dependence between $X$ and $Y$ the easier it is for DAVT-Swap to reject the null.}
    \label{fig:blob-dependent}
\end{figure}

In the main paper, we compared two different two-sample test operators and we established that when the samples $X$ and $Y$ are independent using the swap operator is not very beneficial for the task. In this experiment, we demonstrate the increasing power of DAVT-Swap as soon as the X and Y become more dependent. We revisit the Blob experiment where we model $X_1$ and $Y_1$ (the first dimensions of the dataset) to be dependent and have $\operatorname{corr}(X,Y)=\rho$. We conduct experiments for $\rho=0.1, 0.2, 1$ and we visualize the results in \Cref{fig:blob-dependent}. We can infer that it is easier for DAVT-Swap to reject null the stronger the dependency between $X$ and $Y$.
\subsection{Adversarial Robustness}
\label{app:cifar}
The logit output of the trained ResNet50 is an input to a DNN model with the architecture shown in Table~\ref{tab:cifar}. It is similar to the one of the Blob data with the difference that the hidden layer size is 32. We trained the models for a maximum of 1000 epochs with a learning rate 0.0005 with early stopping with patience 5. We normalize the images at the time of loading.

\subsection{Rotated MNIST}
\label{app:mnist}
The DNN network used here is given in Table \ref{tab:mnist}. The images are normalized before feeding into the network. We traine with $l_1$ ($\lambda_1=0.01$) and $l_2$ ($\lambda_1=0.01$) regularization on the weights. As before, our training procedure is with early stopping with patience 5, learning rate 0.0005, and maximum number of epochs 1000.

\subsection{Conditional Independence Testing under Model-X assumption}
\label{app:cit}
The trained DNN models follow the network architecture in Table~\ref{tab:cit}. The network includes two linear layers with a dropout(p=0.3) layer and a ReLU activation function between them. All models are trained for up to 500 epochs using early stopping based on validation set loss with a patience of 10 and a learning rate of 0.0005.

\begin{table}[!ht]
    \centering
    \begin{tabular}{c|c}
Layer (type)        &       Output Shape   \\ \hline
            Linear-1    &                $[\text{batch size},30]$   \\
       LayerNorm-2    &                 $[\text{batch size},30]$  \\
              ReLU-3    &                $[\text{batch size},30]$   \\
            Linear-4    &                 $[\text{batch size},30]$   \\
       LayerNorm-5    &                 $[\text{batch size},30]$  \\
              ReLU-6   &                 $[\text{batch size},30]$   \\
            Linear-7    &                  $[\text{batch size},\text{output size}]$     \\
    \end{tabular}
    \caption{The network architecture employed in the Blob experiments for all baselines with output size =  2 for S-C2ST and E-C2ST.}
    \label{tab:blob}
\end{table}

\begin{table}[!h]
    \centering
    \begin{tabular}{c|c}
Layer (type)        &       Output Shape   \\ \hline
            Linear-1    &                $[\text{batch size},32]$   \\
       LayerNorm-2    &                 $[\text{batch size},32]$  \\
              ReLU-3    &                $[\text{batch size},32]$   \\
            Linear-4    &                 $[\text{batch size},32]$   \\
       LayerNorm-5    &                 $[\text{batch size},32]$  \\
              ReLU-6   &                 $[\text{batch size},32]$   \\
            Linear-7    &                  $[\text{batch size},\text{output size}]$     \\
    \end{tabular}
    \caption{The network architecture employed in the CIFAR-10 experiments for all baselines with output size = 2 for S-C2ST and E-C2ST.}
    \label{tab:cifar}
\end{table}
\begin{table}[!h]
    \centering
    \begin{tabular}{c|c}
Layer (type)        &       Output Shape   \\ \hline
            Linear-1    &                $[\text{batch size},128]$   \\
              ReLU-2    &                $[\text{batch size},128]$   \\
            Dropout(p=0.5)-3    &                $[\text{batch size},128]$   \\
            Linear-4    &                 $[\text{batch size},64]$   \\
              ReLU-5   &                 $[\text{batch size},64]$   \\
               Dropout(p=0.5)-6    &                $[\text{batch size},64]$   \\
            Linear-7    &                  $[\text{batch size},\text{output size}]$     \\
    \end{tabular}
    \caption{The network architecture employed in the MNIST experiments.}
    \label{tab:mnist}
\end{table}
\begin{table}[!h]
    \centering
    \begin{tabular}{c|c}
Layer (type)        &       Output Shape   \\ \hline
            Linear-1    &                $[\text{batch size},128]$   \\
              ReLU-2    &                $[\text{batch size},
            128]$   \\
            Dropout(p=0.3)-3    &                $[\text{batch size},128]$  \\
             Linear-4    &                $[\text{batch size},\text{output size}]$ 
    \end{tabular}
    \caption{The network architecture employed in the CIT experiments.}
    \label{tab:cit}
\end{table}
\newpage
\section{Practical Considerations}

\begin{wrapfigure}[19]{r}{0.55\textwidth}
    \centering
\includegraphics[scale=0.7]{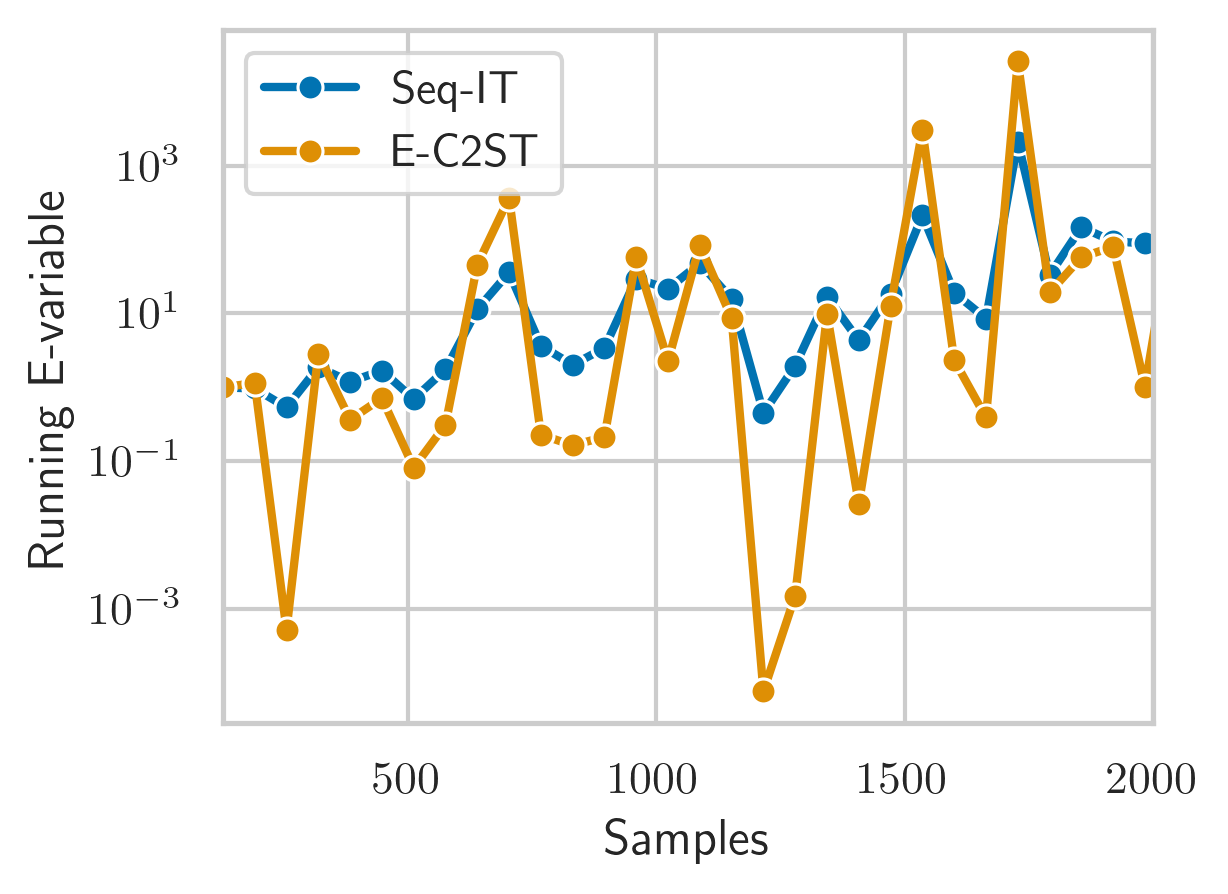}
    \caption{Running E-variables for Seq-IT and E-C2ST. ONS betting strategy decreases the variance of E-C2ST. In other words, unlike the E-C2ST bettor,  the Seq-IT one does not place the entire wealth at stage $t$.}
    \label{fig:runninge}
\end{wrapfigure}

\paragraph{ONS reduces Variance.}
In this example, we want to illustrate the effect of the betting strategy with ONS on the running conditional E-variable. Recall that Seq-IT builds on E-C2ST by incorporating a betting strategy for calculating the test statistic. As a result, the computed Seq-IT conditional E-variable has a lower variance. This is visualized in \Cref{fig:runninge}. In this experiment, we show the running conditional Seq-IT and E-C2ST E-variables over time for the CIFAR-10 experiment. While the E-C2ST E-variable changes abruptly over time, Seq-IT seems to stabilize this statistic and reduce the jumps in the E-variable.

   \paragraph{The model training and architecture.} 
    \label{subsec:fine-tuning}

   Different architectures, from convolutional to recurrent models, can affect the performance of all data-driven tests, a discussion alluded to in \Cref{subsub:low}. Therefore, appropriate DNN selection becomes critical to ensure the reliability and robustness of conclusions derived from these tests.  While all the considered sequential data-driven statistical tests have a finite-sample Type I error control, unlike most classical testing procedures, researchers must be familiar with the specific requirements of their statistical tests in order to select a DNN that maximizes power for a given sample size.
           
    There is another dimension of DNN training worth exploring. Unlike traditional batch mode training, online training is designed for continuous data streams, allowing DNNs to adapt and evolve in real time. This procedure allows to disregard previous data batches and to update the DNN only from a single batch. This can make sequential testing methods even more appealing for a wide range of applications.

    \paragraph{Alternative computation of the betting score.} 
        In~\Cref{sec:proposed-approach}, we defined the betting score $S_t$ in~\eqref{eq:wealth-increment} by taking the product of $(1 + \gtilde_{\theta_{t-1}}(Z))$ over all $Z$ in the mini-batch $B_t$. An alternative, and equally valid, way of defining the wealth process is by taking the average; that is, 
        \begin{align}
            \label{eq:average}
            S_t = \frac{1}{b} \lp  \sum_{Z \in B_t} 1 + \gtilde_{\theta_{t-1}}(Z) \rp = 1 + \frac{1}{b} \sum_{Z \in B_t}\gtilde_{\theta_{t-1}}(Z). 
        \end{align}
        The resulting wealth process has smaller variance than our proposal but this comes at the cost of power for very large mini-batches. This suggests that the optimal betting score construction should lie somewhere in between these two extremes to achieve the best bias variance trade-off.  For example, at test time, a mini-batch can be partitioned into sufficiently small ones for which we compute the increments using the new proposed updating scheme. A thorough exploration of the design of optimal betting scores is an interesting question for future work.

    \paragraph{Unpaired Data.} In the main paper, we mostly presented scenarios where the samples of $X_t$ and $Y_t$ are observed simultaneously. However, our framework is not only limited to paired data. It can also be applied in a more general setting when this assumption does not hold. 
    
    This scenario has been discussed in \cite{shekhar2023nonparametric}. There, the authors propose to use betting scores that align with the proposal in the previous point. More precisely, consider a two-sample test for batches $B_t = \{X_{tb+j}\}_{j=1}^{b_{t1}}\cup \{Y_{tb+j}\}_{j=1}^{b_{t2}}$ consisting of $b_{t1}+b_{t2}$ observations of $X$ and $Y$. Then, we can define the increments as 
    \begin{align}
        S_t = 1+\frac{1}{b_{t1}}\sum_{j=1}^{b_{t1}}g_{\theta_{t-1}}(X_{tb+j})-\frac{1}{b_{t2}}\sum_{j=1}^{b_{t2}}g_{\theta_{t-1}}(Y_{tb+j})
    \end{align}
    or
    \begin{align}
        S_t = 1+\sigma(\sum_{j=1}^{b_{t1}}g_{\theta_{t-1}}(X_{tb+j})-\sum_{j=1}^{b_{t2}}g_{\theta_{t-1}}(Y_{tb+j}))
    \end{align}
    where $\sigma:\mathbb{R}\to[-1,1]$ is an monotone increasing function with $\sigma(-x) = -\sigma(x)$. While this does not quite match our framework, it is a way to model unpaired data.
    
    We can fit this problem into our formalism by considering random variable $L\in\{-1,1\}$ and $W$ such that $P(W|L=1)=P(X)$ and $P(W|L=-1)=P(Y)$. We can test whether $W$ and $L$ are independent instead of considering the classical two-sample test. If the samples $X_t$ and $Y_t$ are somewhat dependent, we can apply averaging to obtain the increments (see \eqref{eq:average}); otherwise, we can stick to our proposal.


\end{document}